\documentclass{clv2}
\usepackage{amsmath}
\usepackage{amssymb}
\usepackage{multirow}
\usepackage[caption=false]{subfig}
\usepackage{algorithm}
\usepackage[noend]{algpseudocode}

\algnewcommand{\algorithmicgoto}{\textbf{goto}}
\algnewcommand{\Goto}[1]{\algorithmicgoto~\ref{#1}}
\newcommand*\Let[2]{\State #1 $\gets$ #2}

\newcommand*\AddOne[1]{\State #1 $++$}

\newcommand{\boxnum}[1]{{\setlength{\fboxsep}{0.8pt}\raisebox{0.8pt}{\hspace{1pt}\fbox{\tiny #1}\hspace{1pt}}}}
\newcommand{\ind}[1]{\ensuremath{_{\kern-0.8pt\boxnum{#1}}}}
\newcommand{\tabincell}[2]{\begin{tabular}{@{}#1@{}}#2\end{tabular}}
\newcolumntype{L}[1]{>{\raggedright\let\newline\\\arraybackslash\hspace{0pt}}m{#1}}

\DeclareMathOperator*{\argmax}{arg\,max}
\DeclareMathOperator\beamsearch{\textsc{BeamSeach}}

\issue{$\times$}{$\times$}{2013}

\runningtitle{A Statistical Parsing Framework for Sentiment Classification}

\runningauthor{Li Dong, Furu Wei, et al.}

\begin{document}

\title{A Statistical Parsing Framework for Sentiment Classification}

\author{Li Dong\thanks{State Key Laboratory of Software Development Environment, Beihang University, XueYuan Road No.37, HaiDian District, Beijing, P.R. China 100191. E-mail: donglixp@gmail.com; kexu@nlsde.buaa.edu.cn} \thanks{Contribution during internship at Microsoft Research.}}
\affil{Beihang University}

\author{Furu Wei\thanks{Natural Language Computing group, Microsoft Research Asia, Building 2, No. 5 Danling Street, Haidian District, Beijing, P.R. China 100080. E-mail: \{fuwei, shujliu, mingzhou\}@microsoft.com.} \thanks{Corresponding Author.}}
\affil{Microsoft Research}

\author{Shujie Liu$^{\dag}$}
\affil{Microsoft Research}

\author{Ming Zhou$^{\dag}$}
\affil{Microsoft Research}

\author{Ke Xu$^{\ast}$}
\affil{Beihang University}


\maketitle

\begin{abstract}
We present a statistical parsing framework for sentence-level sentiment classification in this article.
Unlike previous works that employ syntactic parsing results for sentiment analysis, we develop a statistical parser to directly analyze the sentiment structure of a sentence. 
We show that complicated phenomena in sentiment analysis (e.g., negation, intensification, and contrast) can be handled the same as simple and straightforward sentiment expressions in a unified and probabilistic way.
We formulate the sentiment grammar upon Context-Free Grammars (CFGs), and provide a formal description of the sentiment parsing framework. We develop the parsing model to obtain possible sentiment parse trees for a sentence, from which the polarity model is proposed to derive the sentiment strength and polarity, and the ranking model is dedicated to selecting the best sentiment tree.
We train the parser directly from examples of sentences annotated only with sentiment polarity labels but without any syntactic annotations or polarity annotations of constituents within sentences.
Therefore we can obtain training data easily. In particular, we train a sentiment parser, s.parser, from a large amount of review sentences with users' ratings as rough sentiment polarity labels.
Extensive experiments on existing benchmark datasets show significant improvements over baseline sentiment classification approaches.
\end{abstract}

\newpage
\section{Introduction}
Sentiment analysis~\cite{pangbosurvey2008,bingliubook2012} has received much attention from both research and industry communities in recent years. Sentiment classification, which identifies sentiment polarity (positive or negative) from text (sentence or document), has been the most extensively studied task in sentiment analysis. Up until now, there have been two mainstream approaches for sentiment classification. The lexicon-based approach~\cite{TurneyACL2002,cl:lexicon} aims to aggregate the sentiment polarity of a sentence from the polarity of words or phrases found in the sentence, while the learning-based approach~\cite{pangsvm2002} treats sentiment polarity identification as a special text classification task and focuses on building classifiers from a set of sentences (or documents) annotated with their corresponding sentiment polarity.

The lexicon-based sentiment classification approach is simple and interpretable, but suffers from scalability and is inevitably limited by sentiment lexicons that are commonly created manually by experts. It has been widely recognized that sentiment expressions are colloquial and evolve over time very frequently. Taking tweets from Twitter\footnote{http://twitter.com} and movie reviews in IMDB\footnote{http://www.imdb.com} as examples, people use very casual language as well as informal and new vocabulary to comment on general topics and movies. In practice, it is not feasible to create and maintain sentiment lexicons to capture sentiment expressions with high coverage. On the other hand, the learning-based approach relies on large annotated samples to overcome the vocabulary coverage and deals with variations of words in sentences. Human ratings in reviews~\cite{MaasACL2011} and emoticons in tweets~\cite{DavidovEmoticonColing2010,moodlens} are extensively used to collect a large number of training corpora to train the sentiment classifier. However, it is usually not easy to design effective features to build the classifier. Among others, unigrams have been reported as the most effective features~\cite{pangsvm2002} in sentiment classification. 

Handling complicated expressions delivering people's opinions is one of the most challenging problems in sentiment analysis. Among others, compositionalities such as negation, intensification, contrast, and their combinations are typical cases. We show some concrete examples below.

\begin{exlist}
\item The movie is \underline{not} \textit{good}. [negation]
\item The movie is \underline{very} \textit{good}. [intensification]
\item The movie is \underline{not} \textit{funny} \underline{at all}. [negation + intensification]
\item The movie is \textit{just so so}, \underline{but} i still \textit{like} it. [contrast]
\item The movie is \underline{not} \underline{very} \textit{good}, \underline{but} i still \textit{like} it. [negation + intensification + contrast]
\end{exlist}

The negation expressions, intensification modifiers, and the contrastive conjunction can change the polarity ((1), (3), (4), (5)), strength ((2), (3), (5)), or both ((3), (5)) of the sentiment of the sentences. We do not need any detailed explanations here as they can be commonly found and easily understood in people's daily lives. Existing works to address these issues usually relies on syntactic parsing results either used as features~\cite{Choi:2008:compostional,Moilanen:Packed2010} in learning based methods or hand-crafted rules~\cite{Moilanen:Sentiment:Composition,conf/cikm/JiaYM09,neg:jingjing,klenner-petrakis-fahrni:2009:RANLP09} in lexicon based methods. However, even with the difficulty and feasibility of deriving the sentiment structure from syntactic parsing results put aside, it is an even more challenging task to generate stable and reliable parsing results for text that is ungrammatical in nature and has a high ratio of out-of-vocabulary words. The accuracy of the linguistic parsers trained on standard datasets (e.g., the Penn Treebank~\cite{penn:treebank:Marcus:1993}) drops dramatically on user-generated-content (e.g., reviews, tweets, etc.), which is actually the prime focus of sentiment analysis algorithms. The error, unfortunately, will propagate downstream in the process of sentiment analysis methods building upon parsing results.

We therefore propose directly analyzing the sentiment structure of a sentence. The nested structure of sentiment expressions can be naturally modeled in a similar fashion as statistical syntactic parsing, which aims to find the linguistic structure of a sentence.
This idea creates many opportunities for developing sentiment classifiers from a new perspective.
The most challenging problem and barrier in building a statistical sentiment parser lies in the acquisition of training data. Ideally, we need examples of sentences annotated with polarity for the whole sentence as well as sentiment tags for constituents within a sentence, as with the Penn TreeBank for training traditional linguistic parsers. However, this is not practical as the annotations will be inevitably time consuming and require laborious human efforts.
Therefore, it is better to learn the sentiment parser only employing examples annotated with polarity label of the whole sentence. For example, we can collect a huge number of publicly available reviews and rating scores on the web.
People may use ``\textit{the movie is gud}'' (``gud'' is a popular informal expression of ``good'') to express a positive opinion towards a movie, and ``\textit{not a fan}'' to express a negative opinion. Also, we can find review sentences such as ``\textit{The movie is gud, but I am still not a fan.}'' to indicate a negative opinion.
We can then use these two fragments and the overall negative opinion of the sentence to deduce sentiment rules automatically from data. These sentiment fragments (namely dictionary) and rules can be used to analyze the sentiment structure for new sentences.

In this article, we propose a statistical parsing framework to directly analyze the structure of a sentence from the perspective of sentiment analysis. Specifically, we formulate a Context-Free Grammar (CFG) based sentiment grammar.
We then develop a statistical parser to derive the sentiment structure of a sentence. We leverage the CYK algorithm~\cite{Cocke:1969:PLC:1097042,Younger1967189,Kasami:1965:CYK} to conduct bottom-up parsing, and use dynamic programming to accelerate computation.
Meanwhile, we propose the polarity model to derive sentiment strength and polarity of a sentiment parse tree, and the ranking model to select the best one from the sentiment parsing results.
We train the parser directly from examples of sentences annotated with sentiment polarity labels instead of syntactic annotations and polarity annotations of constituents within sentences. Therefore we can obtain training data easily. In particular, we train a sentiment parser, named \textbf{s.parser}, from a large amount of review sentences with users' ratings as rough sentiment polarity labels.
%
The statistical parsing based approach builds a principled and scalable framework to support the sentiment composition and inference which cannot be well handled by bag-of-words approaches.
We show that complicated phenomena in sentiment analysis (e.g., negation, intensification, and contrast) can be handled the same as simple and straightforward sentiment expressions in a unified and probabilistic way.

The major contributions of the work presented in this article are as follows,
\begin{itemize}
\item We propose a statistical parsing framework for sentiment analysis, which is capable of analyzing the sentiment structure for a sentence. This framework can naturally handle compositionality in a probabilistic way. It can be trained from sentences annotated with only sentiment polarity but without any syntactic annotations or polarity annotations of constituents within sentences;
\item We present the parsing model, polarity model, and ranking model in the proposed framework, which are formulated and can be improved independently. It provides a principled and flexible approach to sentiment classification;
\item We implement the statistical sentiment parsing framework, and conduct experiments on several benchmark datasets. The experimental results show that the proposed framework and algorithm can significantly outperform baseline methods.
\end{itemize}

The remainder of this article is organized as follows. We introduce related work in Section~\ref{sec:related_work}. We present the statistical sentiment parsing framework, including the parsing model, polarity model, and ranking model in Section~\ref{sec:sentiment_parsing}. Learning methods for our model are explained in Section~\ref{sec:model_learning}. Experimental results are reported in Section~\ref{sec:experiment}. We conclude this article with future work in Section~\ref{sec:conclusion}.

\section{Related Work} \label{sec:related_work}
In this section, we give a brief introduction to related work about sentiment classification (Section~\ref{rel:sentiment}) and parsing (Section~\ref{rel:parsing}). We tackle the sentiment classification problem in a parsing manner, which is a significant departure from most previous research.

\subsection{Sentiment Classification} \label{rel:sentiment}
Sentiment classification has been extensively studied in the past few years. In terms of text granularity, existing works can be divided into phrase-level, sentence-level or document-level sentiment classification. We focus on sentence-level sentiment classification in this article. Regardless of what granularity the task is performed on, existing approaches deriving sentiment polarity from text fall into two major categories, namely lexicon-based and learning-based approaches.

The lexicon-based sentiment analysis employs dictionary matching on a predefined sentiment lexicon to derive sentiment polarity. These methods often use a set of manually defined rules to deal with the negation of polarity.
\namecite{TurneyACL2002} proposed using the average sentiment orientation of phrases, which contains adjectives or adverbs, in a review to predict its sentiment orientation.
\namecite{Yu+Hatzivassiloglou:03a} calculated a modified log-likelihood ratio for every word by the co-occurrences with positive and negative seed words. To determine the polarity of a sentence, they compare the average log-likelihood value with threshold.
\namecite{cl:lexicon} presented a lexicon-based approach for extracting sentiment from text. They used dictionaries of words with annotated sentiment orientation (polarity and strength) while incorporating intensification and negation.
The lexicon-based methods often achieve high precisions and do not need any labeled samples. But they suffer from coverage and domain adaption problems. Moreover, lexicons are often built and used without considering the context~\cite{WilsonCL2009}. Also, hand-crafted rules are often matched heuristically.

The sentiment dictionaries used for lexicon-based sentiment analysis can be created manually, or automatically using seed words to expand the list of words.
\namecite{Kamps2004,icwsm:WilliamsA09} used various lexical relations (such as synonym and antonym relations) in WordNet to expend a set of seed words.
Some other methods learn lexicons from data directly.
\namecite{Hatzivassiloglou:1997:PSO:976909.979640} used a log-linear regression model with conjunction constraints to predict whether conjoined adjectives have similar or different polarities. Combining conjunction constraints across many adjectives, a clustering algorithm separated the adjectives into groups of different polarity. Finally, adjectives were labeled as positive or negative.
\namecite{Velikovich:2010:VWP:1857999.1858118} constructed a term similarity graph using the cosine similarity of context vectors. They performed graph propagation from seeds on the graph, obtaining polarity words and phrases.
\namecite{Takamura:2005:ESO:1219840.1219857} regarded the polarity of word as spins of electrons, using the mean field approximation to compute the approximate probability function of the system instead of the intractable actual probability function.
\namecite{Kanayama:2006:FAL:1610075.1610125} employed tendencies for similar polarities to appear successively in contexts. They defined density and precision of coherency to filter neutral phrases and uncertain candidates.
\namecite{lexion:ipl:06,lexion:ipl:11} transformed the lexicon learning to an optimization problem, and employed integer linear programming to solve it.
\namecite{Kaji07buildinglexicon} defined Chi-square based polarity value and PMI based polarity value as a polarity strength to filter neutral phrases.
\namecite{deMarneffe:2010:GPL:1858681.1858699} utilized review data to define polarity strength as the expected rating value.
\namecite{Mudinas:2012:CLL:2346676.2346681} used word count as a feature template and trained a classifier using Support Vector Machines with linear kernel. They then regarded the weights as polarity strengths.
\namecite{Krestel:2013:GCS:2481492.2481506} generated topic-dependent lexicons from review articles by incorporating topic and rating probabilities and defined the polarity strength based on the results.
In this article, the lexical relations defined in WordNet are not employed due to its coverage.
Furthermore, most of these methods define different criteria to propagate polarity information of seeds, or employ optimization algorithms and sentence-level sentiment labels to learn polarity strength values. Their goal is to balance the precision and recall of learned lexicons.
We also learn the polarity strength values of phrases from data. However, our primary objective is to obtain correct sentence-level polarity labels, and use them to form the sentiment grammar.

Learning-based sentiment analysis employs machine learning methods to classify sentences or documents into two (negative and positive) or three (negative, positive and neutral) classes.
Previous research has shown that sentiment classification is more difficult than traditional topic-based text classification, although the fact that the number of classes in sentiment classification is smaller than that in topic-based text classification~\cite{pangbosurvey2008}.
\namecite{pangsvm2002} investigated three machine learning methods to produce automated classifiers to generate class labels for movie reviews. They tested them on Na\"{i}ve Bayes, Maximum Entropy, and Support Vector Machine, and evaluated the contribution of different features including unigrams, bigrams, adjectives, and part-of-speech tags. Their experimental results suggested that a SVM classifier with unigram presence features outperforms other competitors.
\namecite{PangACL2004} separated subjective portions from the objective by finding minimum cuts in graphs to achieve better sentiment classification performance.
\namecite{Matsumoto:2005:SCU:2140831.2140873} used text mining techniques to extract frequent subsequences and dependency subtrees, and used them as features of SVM.
\namecite{Mcdonald07structuredmodels} investigated a global structured model for jointly classifying polarity at different levels of granularity. This model allowed classification decisions from one level in the text to influence decisions at another.
\namecite{yessenalina2010multi} used sentence-level latent variables to improve document-level prediction.
\namecite{tackstrom2011discovering} presented a latent variable model for only using document-level annotations to learn sentence-level sentiment labels, and \namecite{Tackstrom:2011} improved it by using a semi-supervised latent variable model to utilize manually crafted sentence labels.
\namecite{Agarwal:2011:SAT:2021109.2021114,Tu:2012:IHS:2390665.2390742} explored part-of-speech tag features and tree-kernel.
\namecite{sidaw12acls} used Support Vector Machine (SVM) built over Na\"{i}ve Bayes log-count ratios as feature values to classify polarity. They showed that SVM was better at full-length reviews, and Multinomial Na\"{i}ve Bayes was better at short-length reviews.
\namecite{DBLP:conf/coling/LiuAG12} proposed a set of heuristic rules based on dependency structure to detect negations and sentiment-bearing expressions.
Most of the above methods are built on bag-of-words features, and sentiment compositions are handled by manually crafted rules. In contrast to these models, we derive polarity labels from tree structures parsed by the sentiment grammar.

There have been several attempts to assume the problem of sentiment analysis is compositional. Sentiment classification can be solved by deriving the sentiment of a complex constituent (sentence) from the sentiment of small units (words and phrases)~\cite{Moilanen:Sentiment:Composition,tree-crf,klenner-petrakis-fahrni:2009:RANLP09,Choi:2010:HSL:1858842.1858892}.
\namecite{Moilanen:Sentiment:Composition} proposed using delicate written linguistic patterns as heuristic decision rules when computing the sentiment from individual words to phrases and finally to the sentence. The manually-compiled rules were powerful to discriminate between the different sentiments in ``effective remedies'' (positive) / ``effective torture'' (negative), and in ``too colorful'' (negative) and ``too sad'' (negative).
\namecite{tree-crf} leveraged a conditional random field model to calculate the sentiment of all the parsed elements in the dependency tree and then generated the overall sentiment. It had an advantage over the rule-based approach~\cite{Moilanen:Sentiment:Composition} in that it did not explicitly denote any sentiment designation to words or phrases in parse trees, Instead, it modeled their sentiment polarity as latent variables with a certain probability of being positive or negative.
\namecite{Councill:2010} used a conditional random field model informed by a dependency parser to detect the scope of negation for sentiment analysis.
Some other methods model sentiment compositionality in the vector space. They regard the composition operator as a matrix, and use matrix-vector multiplication to obtain the transformed vector representation.
\namecite{SocherEtAl2012:MVRNN} proposed a recursive neural network model that learned compositional vector representations for phrases and sentences. Their model assigned a vector and a matrix to every node in a parse tree. The vector captured the inherent meaning of the constituent, while the matrix captured how it changes the meaning of neighboring words or phrases.
\namecite{SocherEtAl2013:RNTN} recently introduced a sentiment treebank based on the results of the Stanford parser~\cite{Klein:2003:AUP:1075096.1075150}. The sentiment treebank included polarity labels of phrases which are annotated by using Amazon Mechanical Turk. The authors trained recursive neural tensor networks on the sentiment treebank. For a new sentence, the model predicted polarity labels based on the syntactic parse tree, and used tensors to handle compositionality in the vector space.
\namecite{dong2014adaptive} proposed employing multiple composition functions in recursive neural models and learn to select them adaptively.
Most previous methods are either rigid in terms of handcrafted rules, or sensitive to the performance of existing syntactic parsers they use. This article addresses sentiment compositions by defining sentiment grammar and borrowing some techniques in the parsing research field. Moreover, our method is in a symbolic way instead of in the vector space.


\subsection{Syntactic Parsing and Semantic Parsing}\label{rel:parsing}
The work presented in this article is close to traditional statistical parsing as we borrow some algorithms to build the sentiment parser.
Syntactic parsers are learned from the Treebank corpora, and find the most likely parse tree with the largest probability.
In this article, we borrow some well-known techniques from syntactic parsing methods~\cite{Charniak:1997:SPC:1867406.1867499,Charniak:2005:CNB:1219840.1219862,McDonald:2005:OLT:1219840.1219852,kubler2009dependency}, such as the CYK algorithm and Context-Free Grammar.
These techniques are used to build the sentiment grammar and parsing model. They provide a natural way to define the structure of sentiment tree and parse sentences to trees.
The key difference lies in that our task is to calculate the polarity label of a sentence, instead of obtaining the parse tree. We only have sentence-polarity pairs as our training instances instead of annotated tree structures. Moreover, in the decoding process, our goal is to compute correct polarity labels by representing sentences as latent sentiment trees.
Recently, \namecite{hall2014less} developed a discriminative constituency parser using rich surface features, adapting it to sentiment analysis.
However, their method relies on phrase-level polarity annotations and syntactic parse results.
Also, only learning interactions between tags and words located at the beginning or the end of spans limits their abilities to process more complex sentiment rules.

Semantic parsing is another body of work related to this article. A semantic parser is used to parse meaning representations for given sentences.
Most existing semantic parsing works~\cite{zelle:aaai96,AAAI136189,Zettlemoyer:2009:LCM:1690219.1690283,kate:acl06,ge:acl06,Zettlemoyer07onlinelearning} relied on fine-grained annotations of target logical forms, which required the supervision of experts and are relatively expensive.
To balance the performance and the amount of human annotation, some works used only question-answer pairs or even binary correct/incorrect signals as their input.
\namecite{Clarke:2010:DSP:1870568.1870571} employed a binary correct/incorrect signal of a database query to map sentences to logical forms. It worked with FunQL language and transformed semantic parsing as an integer linear programming (ILP) problem. In each iteration, it solved ILP and updated the parameters of structural SVM.
\namecite{percyliang} learned a semantic parser from question-answer pairs, where the logical form was modeled as latent tree-based semantic representation.
\namecite{KrishnamurthyEMNLP2012} presented a method for training a semantic parser using a knowledge base and an unlabeled text corpus, without any individually annotated sentences.
\namecite{artzi2013weakly} used various types of weak supervision to learn a grounded Combinatory Categorial Grammar semantic parser, which took context into consideration.
\namecite{nan2014sp} presented a translation-based weakly-supervised semantic parsing method to translate questions to answers based on CYK parsing. A log-linear model is defined to score derivations.
All these weakly supervised semantic parsing methods learned to transform a natural language sentence to its semantic representation without annotated logical form.
In this work, we build a sentiment parser. Specifically, we employ a modified version of the CYK algorithm which parses sentences in a bottom-up fashion. We use the log-linear model to score candidates generated by beam search. Instead of using question-answer pairs, sentence-polarity pairs are used as our weak supervisions. We also employ the parameter estimation algorithm proposed by~\namecite{percyliang}.

\section{Statistical Sentiment Parsing} \label{sec:sentiment_parsing}
\begin{figure}[tb]
\begin{center}
\includegraphics[width=0.97\textwidth]{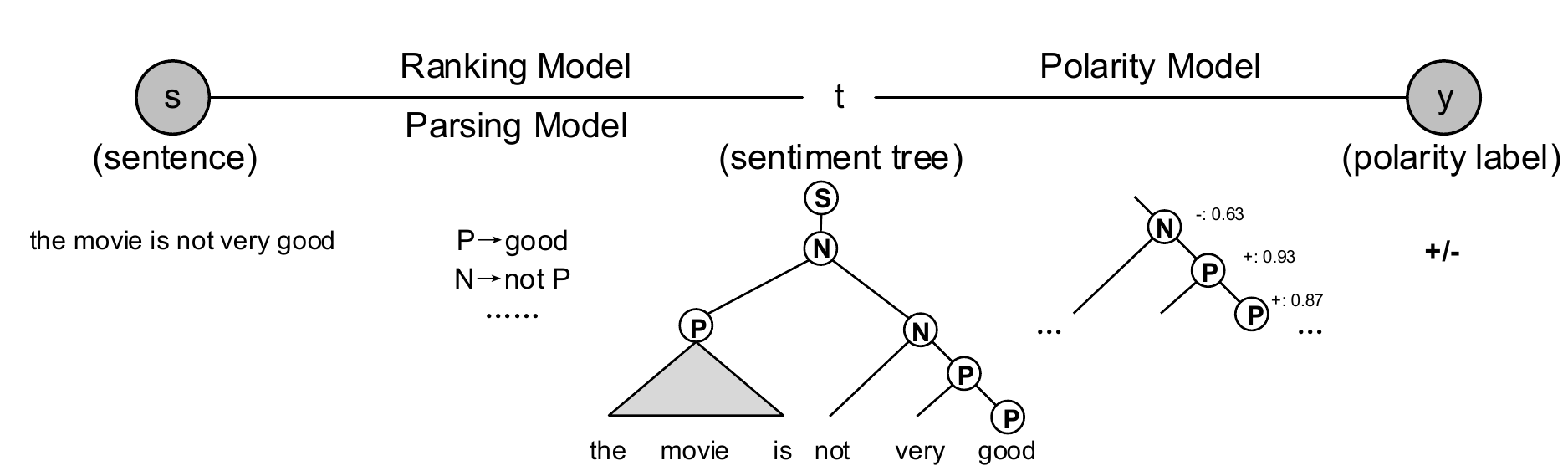}
\caption{The parsing model and ranking model are used to transform the input sentence $s$ to the sentiment tree $t$ with the highest ranking score. Moreover, the polarity model defines how to compute polarity values for the rules of the sentiment grammar. The sentiment tree $t$ is evaluated with respect to the polarity model to produce the polarity label $y$.}
\label{fig:graph_model}
\end{center}
\end{figure}

We present the statistical parsing framework for sentence-level sentiment classification in this section. The underlying idea is to model sentiment classification as a statistical parsing process.
Figure~\ref{fig:graph_model} shows the overview of the statistical sentiment parsing framework. There are three major components. The input sentence $s$ is transformed into and represented by sentiment trees derived from the parsing model (Section~\ref{sec:parsing_model}), using the sentiment grammar defined in Section~\ref{sec:sentiment_grammar}. Trees are scored by the ranking model in Section~\ref{sec:ranking_model}. The sentiment tree with the highest ranking score is treated as the best derivation for $s$.
Furthermore, the polarity model (Section~\ref{sec:polarity_model}) is used to compute polarity values for the sentiment trees.

Notably, the sentiment trees $t$ are unobserved during training. We can only observe the sentence $s$ and its polarity label $y$ in training data. In other words, we train the model directly from the examples of sentences annotated only with sentiment polarity labels but without any syntactic annotations or polarity annotations of the constituents within sentences. To be specific, we first learn the sentiment grammar and the polarity model from data as described in Section~\ref{sec:grammar_learning}.
Then, given the sentence and polarity label pairs $\left( s , y \right)$, we search the latent sentiment trees $t$ and estimate the parameters of the ranking model as detailed in Section~\ref{sec:ranking_model_learning}.

\begin{figure}[b]
\begin{center}
\includegraphics[width=0.83\textwidth]{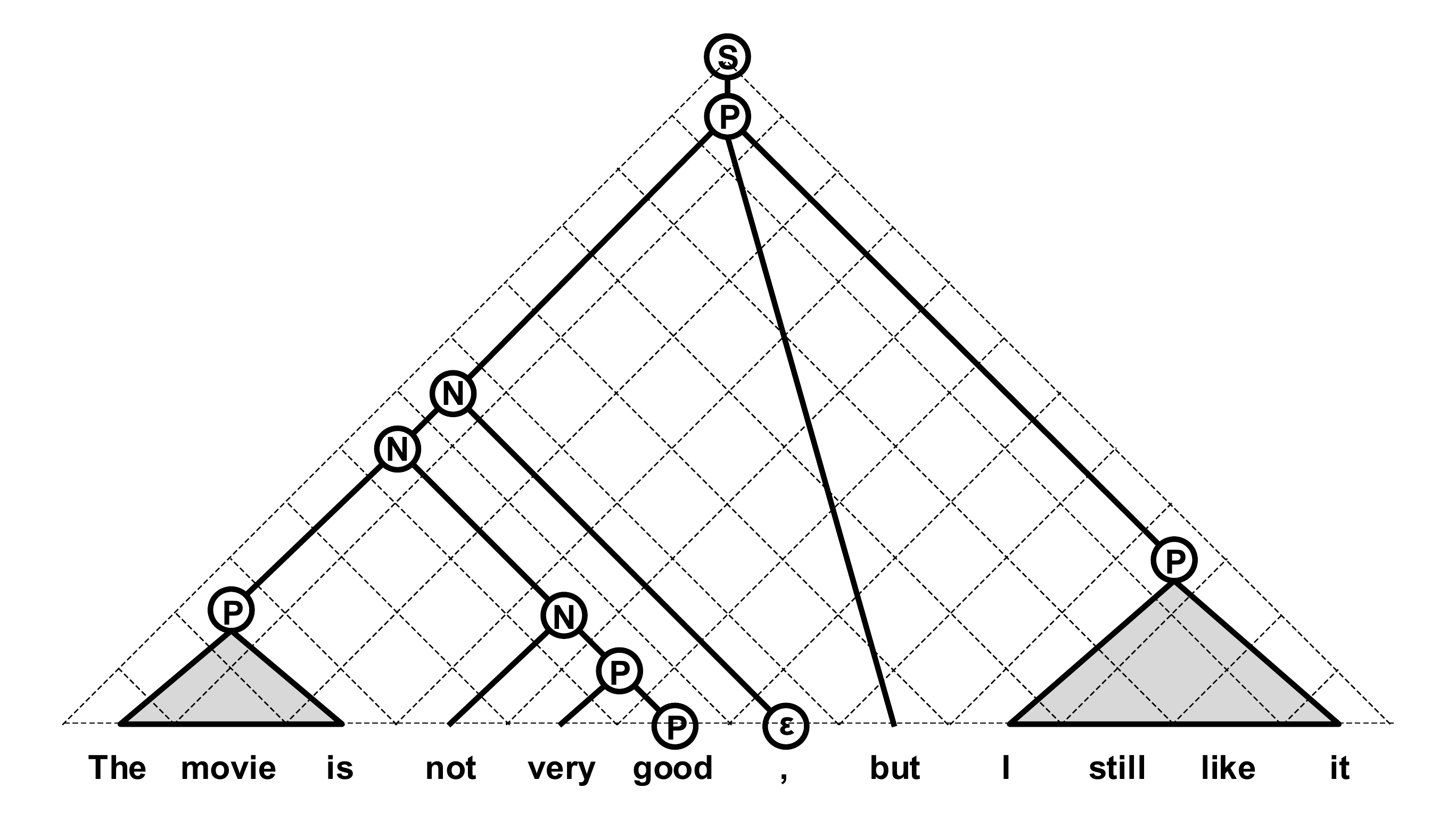}
\caption{Sentiment structure for the sentence ``\emph{The movie is \underline{not} \underline{very} good, \underline{but} i still like it}''. The rules employed in the derivation process include \{${P} \rightarrow \mbox{the~movie~is}$; ${P} \rightarrow \mbox{good}$; ${P} \rightarrow \mbox{i~still~like~it}$; ${P} \rightarrow \mbox{very}~{P}$; ${N} \rightarrow \mbox{not}~{P}$; ${N} \rightarrow {P} {N}$; ${N} \rightarrow {N} \mathcal{E}$;
$\mathcal{E} \rightarrow \mbox{,}$;
${P} \rightarrow {N}~\mbox{but}~{P}$; ${S} \rightarrow {P}$\}.}
\label{fig:example_sentiment_structure}
\end{center}
\end{figure}

To better illustrate the whole process, we describe the sentiment parsing procedure using an example sentence, ``\emph{The movie is \underline{not} \underline{very} good, \underline{but} i still like it}''. The sentiment polarity label of the above sentence is ``positive''.
There is negation, intensification, and contrast in this example, which are difficult to capture using bag-of-words classification methods. This sentence is a complex case that demonstrates the capability of the proposed statistical sentiment parsing framework, which motivates the work in this article.
The statistical sentiment parsing algorithm may generate a number of sentiment trees for the input sentence. Figure~\ref{fig:example_sentiment_structure} shows the best sentiment parse tree.
It shows that the statistical sentiment parsing framework can deal with the compositionality of sentiment in a natural way.
In Table~\ref{table:parsing_process}, we list the sentiment rules used during the parsing process. We show the generation process of the sentiment parse tree from the bottom-up and the calculation of sentiment strength and polarity for every text span in the parsing process.

\begin{table}[tb]
\caption{Parsing process for the sentence ``\emph{The movie is \underline{not} \underline{very} good, \underline{but} i still like it}''. $[i , {Y} , j]$ represents the text spanning from $i$ to $j$ is derived to symbol $Y$. $N$ and $P$ are non-terminals in the sentiment grammar, while $\mathcal{N}$ and $\mathcal{P}$ represent polarities of sentiment.}
\begin{tabular*}{\textwidth}{l l l l}
\hline
Span & Rule & Strength & Polarity \\ \hline
$[0,{P}, 3]$: the movie is& ${P} \rightarrow \mbox{the~movie~is}$ & 0.52 & $\mathcal{P}$ \\
$[5,{P}, 6]$: good & ${P} \rightarrow \mbox{good}$ & 0.87 & $\mathcal{P}$ \\
$[6, \mathcal{E} , 7]$: , & $\mathcal{E} \rightarrow \mbox{,}$ & - & - \\
$[8,{P}, 11]$: i still like it & ${P} \rightarrow \mbox{i~ still~ like~ it}$ & 0.85 & $\mathcal{P}$ \\
$[4,{P}, 6]$: very good & ${P} \rightarrow \mbox{very}~P$ & 0.93 & $\mathcal{P}$ \\
$[3,{N}, 6]$: not very good & ${N} \rightarrow \mbox{not}~P$ & 0.63 & $\mathcal{N}$ \\
$[0,{N}, 6]$: the movie is not very good & ${N} \rightarrow {P} {N}$ & 0.60 & $\mathcal{N}$ \\
$[0,{N}, 7]$: the movie is not very good, & ${N} \rightarrow {N} \mathcal{E}$ & 0.60 & $\mathcal{N}$ \\
$[0,{P}, 11]$: the movie is not very good, but i still like it & ${P} \rightarrow {N} ~\mbox{but}~ P$ & 0.76 & $\mathcal{P}$ \\
$[0, {S}, 11]$: the movie is not very good, but i still like it & ${S} \rightarrow {P}$ & 0.76 & $\mathcal{P}$ \\
\hline
\end{tabular*}
\label{table:parsing_process}
\end{table}

In the following sections, we first provide a formal description of the sentiment grammar in Section~\ref{sec:sentiment_grammar}. We then present the details of the parsing model in Section~\ref{sec:parsing_model}, the ranking model in Section~\ref{sec:ranking_model}, and the polarity model in Section~\ref{sec:polarity_model}.

\subsection{Sentiment Grammar} \label{sec:sentiment_grammar}
We develop the sentiment grammar upon CFG (Context-Free Grammar)~\cite{chomsky56}. Let $\mathcal{G} = < V , \Sigma , {S} , R >$ denote a CFG, where $V$ is a finite set of non-terminals, $\Sigma$ is a finite set of terminals (disjointed from $V$), ${S} \in V$ is the start symbol, and $R$ is a set of rewrite rules (or production rules) of the form $A \rightarrow c$ where $A \in V$ and $c \in {(V \cup \Sigma)}^{*}$.
We use $\mathcal{G}_{s} = < V_s , \Sigma_s , {S} , R_s >$ to denote the sentiment grammar in this article. The non-terminal set is denoted as $V_s = \{{N},{P},{S},{\mathcal{E}} \}$, where $S$ is the start symbol, the non-terminal $N$ represents the negative polarity, and the non-terminal $P$ represents the positive polarity. The rules in $R_s$ are divided into the following six categories:
\begin{itemize}
\item \textit{Dictionary rules}: ${X} \rightarrow w_0^k$, where ${X} \in \{{N},{P}\}$, $w_0^k = w_0 \dots w_{k-1}$, and $w_0^k \in \Sigma_s^{+}$. These rules can be regarded as the sentiment dictionary used in traditional approaches. They are basic sentiment units assigned with polarity probabilities. For instance, ${P} \rightarrow \mbox{good}$ is a dictionary rule;
\item \textit{Combination rules}: ${X} \rightarrow c$, where $c \in (V_s \cup \Sigma_s)^{+}$, and two successive non-terminals are not allowed. There is at least one terminal in $c$. These rules combine terminals and non-terminals, such as ${N} \rightarrow \mbox{not}~P$, and ${P} \rightarrow {N} ~\mbox{but}~ P$. They are used to handle negation, intensification, and contrast in sentiment analysis. The number of non-terminals in a combination rule is restricted to one and two; 
\item \textit{Glue rules}: ${X} \rightarrow {X}_{1} {X}_{2}$, where ${X} , {X}_{1} , {X}_{2} \in \{{N},{P}\}$. These rules combine two text spans which are derived into ${X}_{1}$ and ${X}_{2}$, respectively;
\item \textit{OOV rules}: $\mathcal{E} \rightarrow w_0^k$, where $w_0^k \in \Sigma^{+}$. We use these rules to handle Out-Of-Vocabulary (OOV) text spans whose polarity probabilities are not learned from data;
\item \textit{Auxiliary rules}: ${X} \rightarrow \mathcal{E} {X}_{1}$, ${X} \rightarrow {X}_{1} \mathcal{E}$, where ${X} , {X}_{1} \in \{{N},{P}\}$. These rules combine a text span with polarity and an OOV text span;
\item \textit{Start rules}: ${S} \rightarrow Y$, where $Y \in \{{N},{P} , \mathcal{E} \}$. The derivations begin with ${S}$, and ${S}$ can be derived to $N$, $P$, and $\mathcal{E}$.
\end{itemize}

Here, $X$ represents the non-terminals $N$ or $P$.
The dictionary rules and combinations rules are automatically extracted from the data. We will describe the details in Section~\ref{sec:grammar_learning}. By employing these rules, we can derive the polarity label of a sentence from the bottom-up.
The glue rules are used to combine polarity information of two text spans together, and it treats the combined parts as independent.
In order to tackle the Out-Of-Vocabulary (OOV) problem, we treat a text span that consists of OOV words as empty text span, and derive them to $\mathcal{E}$. The OOV text spans are combined with other text spans without considering their sentiment information. Finally, each sentence is derived to the symbol $S$ using the start rules which are the beginnings of derivations. We can use the sentiment grammar to compactly describe the derivation process of a sentence.

\subsection{Parsing Model} \label{sec:parsing_model}
We present the formal description of the statistical sentiment parsing model following deductive proof systems~\cite{Shieber:1995:PID,Goodman:1999:SP:973226.973230} as used in traditional syntactic parsing. For a concrete example,
\begin{equation}
\label{eq:weight_cky_binary_rule}
\frac { ( {A}\rightarrow {B}{C} ) \quad [i , {B} , k]  \quad [k, {C} , j]}{[i , {A} , j] }
\end{equation}
which represents if we have the rule ${A}\rightarrow {B}{C}$ and ${B} \overset{*}{\Rightarrow} w_{i}^{k}$ and ${C} \overset{*}{\Rightarrow} w_{k}^{j}$ ($\overset{*}{\Rightarrow}$ is used to represent the reflexive and transitive closure of immediate derivation), then we can obtain ${A} \overset{*}{\Rightarrow} w_{i}^{j}$.
By adding a unary rule
\begin{equation}
\label{eq:weight_cky_unary_rule}
\frac { ( {A}\rightarrow w^j_i ) }{ [i,{A},j] }
\end{equation}
with the binary rule in Equation~\eqref{eq:weight_cky_binary_rule}, we can express the standard CYK algorithm for CFG in Chomsky Normal Form (CNF). And the goal is $[0,{S},n]$, in which ${S}$ is the start symbol and $n$ is the length of the input sentence.
In the above CYK example, the \textbf{term} in deductive rules can be one of the following two forms:
\begin{itemize}
\item $[i , X , j]$ is an \textbf{item} representing a subtree rooted in $X$ spanning from $i$ to $j$, or
\item $\left( X \rightarrow \gamma \right)$ is a \textbf{rule} in the grammar.
\end{itemize}

Generally, we represent the form of an inference rule as:
\begin{equation}
\label{eq:rule_basic_form}
\frac { (r) \quad {H}_1  \quad \dots \quad {H}_K }{ [i , X , j] }
\end{equation}
where if all the terms $r$ and ${H}_k$ are true, then we can infer $[i , X , j]$ as true. Here, $r$ denotes a sentiment rule, and ${H}_k$ denotes an item. When we refer to both rules and items, we employ the word \textbf{terms}.

Theoretically, we can convert the sentiment rules to CNF versions, and then employ the CYK algorithm to conduct parsing. Since the maximum number of non-terminal symbols in a rule is already restricted to two, we formulate the statistical sentiment parsing based on a customized CYK algorithm which is similar to the work of~\namecite{Chiang:2007:HPT:1268656.1268659}.
Let ${X} , {X}_{1} , {X}_{2}$ represent the non-terminals $N$ or $P$, the inference rules for the statistical sentiment parsing are summarized in Figure~\ref{fig:model_inference_rule_basic}.

\begin{figure}[t]
\begin{center}
{
\[
\begin{aligned}
&\frac { ({X} \rightarrow w^j_i ) }{ [i , {X} , j] }
\\
&\frac { ({X} \rightarrow {w}^{i_1}_{i} {X}_{1} {w}^{j}_{j_1} ) \quad  [ {i}_{1} ,{X}_{1}  , {j}_{1} ] }{ [i,{X},j]   }
\\
&\frac { ({X}\rightarrow {w}^{i_1}_{i} {X}_{1} {w}^{i_2}_{j_1} {X}_{2} {w}^{j}_{j_2} ) \quad  [ {i}_{1} , {X}_{1} ,{j}_{1}]  \quad [ {i}_{2} ,{X}_{2} , {j}_{2}] }{ [i,{X},j]  }
\\
&\frac { ({X}\rightarrow {X}_{1} {X}_{2} ) \quad [i, {X}_{1} ,k]  \quad [k, {X}_{2} ,j] }{ [i,{X},j]  }
\\
&\frac { (\mathcal{E} \rightarrow w^j_i ) }{ [i , \mathcal{E} , j] }
\\
&\frac { ({X}\rightarrow \mathcal{E} {X}_{1} ) \quad [i, \mathcal{E} ,k]  \quad [k, {X}_{1} ,j] }{ [i,{X},j]  }
\\
&\frac { ({X}\rightarrow {X}_{1} \mathcal{E} ) \quad [i, {X}_{1} ,k]  \quad [k, \mathcal{E} ,j] }{ [i,{X},j]  }
\\
&\mbox{where}~{X} , {X}_{1} , {X}_{2}~\mbox{represent}~{N}~\mbox{or}~{P}.
\end{aligned}
\]
}
\caption{Inference rules for the basic parsing model.}
\label{fig:model_inference_rule_basic}
\end{center}
\end{figure}

\subsection{Ranking Model} \label{sec:ranking_model}
The parsing model generates many candidate parse trees $T(s)$ for a sentence $s$. The goal of the ranking model is to score and rank these parse trees. The sentiment tree with the highest score is treated as the best representation for sentence $s$.
We extract a feature vector $\mathbf{\phi}(s,t) \in \mathcal{R}^{d}$ for the specific sentence-tree pair $(s,t)$, where $t \in T(s)$ is the parse tree. Let $\mathbf{\psi} \in \mathcal{R}^{d}$ be the parameter vector for the features. We use the log-linear model to calculate a probability $p(t|s;T,\mathbf{\psi})$ for each parse tree $t \in T(s)$.
The probabilities indicate how likely the trees are to produce correct predictions.
Given the sentence $s$ and parameters $\mathbf{\psi}$, the log-linear model defines a conditional probability:
\begin{equation}
p(t|s;T,\mathbf{\psi}) = \exp{\{{\mathbf{\phi}(s,t)}^{\mathsf{T}} \mathbf{\psi} - \mathbf{A}(\mathbf{\psi};s,T) \}}
\end{equation}
\begin{equation}
\label{eq:log_linear_model_A}
\mathbf{A}(\mathbf{\psi};s,T) = \log {\sum_{t \in T(s)} {\exp{\{\mathbf{\phi}(s,t)^{\mathsf{T}} \mathbf{\psi} \}}}}
\end{equation}
where $\mathbf{A}(\mathbf{\psi};s,T)$ is the log-partition function with respect to $T(s)$. The log-linear model is a discriminative model, and it is widely used in natural language processing.
We can use ${\mathbf{\phi}(s,t)}^{\mathsf{T}} \mathbf{\psi}$ as the score of the parse tree without normalization in the decoding process, because $p(t|s;T,\mathbf{\psi}) \propto {\mathbf{\phi}(s,t)}^{\mathsf{T}} \mathbf{\psi}$ and this will not change the ranking order.

\subsection{Polarity Model} \label{sec:polarity_model}
The goal of the polarity model is to model the calculation of sentiment strength and polarity of a text span from its sub-spans in the parsing process. It is specified in terms of the rules employed in the parsing process.
We expand the notations in the inference rule~\eqref{eq:rule_basic_form} to incorporate the polarity model.
The new form of inference rule is:
\begin{equation}
\label{eq:rule_general_form_with_computation_model}
\frac { (r) \quad {H}_1 \Phi_1  \quad \dots \quad {H}_K \Phi_K }{ [i , X , j] \Phi }
\end{equation}
in which $r , {H}_1, \dots, {H}_K$ are the terms described in Section~\ref{sec:parsing_model}. Every item ${H}_k$ is assigned polarity strength $\Phi_k : \begin{cases}
P(\mathcal{N}| w_{i_k}^{j_k} ) \\
P(\mathcal{P}| w_{i_k}^{j_k} ) \end{cases}$ for text span $w_{i_k}^{j_k}$.
For the item $[i,{X},j]$, the polarity model $\Phi(r, \Phi_1,\dots,\Phi_K)$ is defined as a function which takes the rule $r$ and polarity strength of sub-spans as input.

The polarity strength obtained by the polarity model should satisfy two constraints. First, the values calculated by the polarity model are non-negative, i.e., $P(\mathcal{X}| w_i^j ) \ge 0 , P(\overline{\mathcal{X}}| w_i^j ) \ge 0$. Second, the positive and negative polarity values are normalized to 1, i.e., $P(\mathcal{X}| w_i^j ) + P(\overline{\mathcal{X}}| w_i^j ) = 1$.
Notably, $\overline{\mathcal{X}} = \begin{cases}
\mathcal{P}, &\mathcal{X}=\mathcal{N} \\
\mathcal{N}, &\mathcal{X}=\mathcal{P} \end{cases}$ is the opposite polarity of $\mathcal{X}$.

The inference rules with the polarity model are formally defined in Figure~\ref{fig:model_inference_rule}.
In the following part, we define the polarity model for the different types of rules.
If the rule is a dictionary rule ${X} \rightarrow w_i^j$, its sentiment strength is obtained as:
\begin{equation}
\Phi :
\begin{cases}
	P(\mathcal{X}| w_i^j ) = \tilde{P}(\mathcal{X}| w_i^j ) \\
	P(\overline{\mathcal{X}}| w_i^j ) = \tilde{P}(\overline{\mathcal{X}}| w_i^j )
\end{cases}
\end{equation}
where $\mathcal{X} \in \{\mathcal{N},\mathcal{P}\}$ denotes the sentiment polarity of the left hand side of the rule, $\overline{\mathcal{X}}$ is the opposite polarity of $\mathcal{X}$,
and $\tilde{P}(\mathcal{X}| w_i^j ), \tilde{P}(\overline{\mathcal{X}}| w_i^j )$ indicate the sentiment polarity values estimated from training data.

\begin{figure}[t]
\begin{center}
{
\[
\begin{aligned}
&\frac { ({X} \rightarrow w^j_i ) }{ [i , {X} , j]
		P(\mathcal{X}|w_i^j) = \tilde{P}(\mathcal{X}|w_i^j)
	}
\\
&\frac { ({X} \rightarrow {w}^{i_1}_{i} {X}_{1} {w}^{j}_{j_1} ) \quad  [ {i}_{1} ,{X}_{1}  , {j}_{1} ] \Phi_1 }{ [i,{X},j]
		P(\mathcal{X}|w_i^j) = h( \mathbf{\theta}_0 + \mathbf{\theta}_1 P(\mathcal{X}_{1} | w_{i_1}^{j_1}) )  }
\\
&\frac {( {X}\rightarrow {w}^{i_1}_{i} {X}_{1} {w}^{i_2}_{j_1} {X}_{2} {w}^{j}_{j_2} )  \quad  [ {i}_{1} , {X}_{1} ,{j}_{1}] \Phi_1  \quad [ {i}_{2} ,{X}_{2} , {j}_{2}] \Phi_2 }{ [i,{X},j]
		P(\mathcal{X}|w_i^j) = h( \mathbf{\theta}_0 + { \mathbf{\theta}_1 P(\mathcal{X}_1 | w_{i_1}^{j_1})} + { \mathbf{\theta}_2 P(\mathcal{X}_2 | w_{i_2}^{j_2})} ) }
\\
&\frac { ({X}\rightarrow {X}_{1} {X}_{2} ) \quad [i, {X}_{1} ,k] \Phi_1  \quad [k, {X}_{2} ,j] \Phi_2 }{ [i,{X},j]
		P(\mathcal{X}|w_i^j) = \frac { P(\mathcal{X}|w_i^k) P(\mathcal{X}|w_k^j) }{ P(\mathcal{X}|w_i^k) P(\mathcal{X}|w_k^j)+ P(\overline{\mathcal{X}}|w_i^k) P(\overline{\mathcal{X}}|w_k^j) } }
\\
&\frac { (\mathcal{E} \rightarrow w^j_i ) }{ [i , \mathcal{E} , j]
	\circ
}
\\
&\frac { ({X}\rightarrow \mathcal{E} {X}_{1} ) \quad [i, \mathcal{E} ,k] \circ  \quad [k, {X}_{1} ,j] \Phi_1 }{ [i,{X},j]
		P(\mathcal{X}|w_i^j) =  P(\mathcal{X}|w_k^j) }
\\
&\frac {( {X}\rightarrow {X}_{1} \mathcal{E} ) \quad [i, {X}_{1} ,k] \Phi_1  \quad [k, \mathcal{E} ,j] \circ }{ [i,{X},j]
		P(\mathcal{X}|w_i^j) = P(\mathcal{X}|w_i^k) }
\\
\mbox{where}~h(x) = &\frac{1}{1+\exp\{ -x \}}~\mbox{is~a~logistic~function,} \circ \mbox{represents~the~absence, and}~{X} , {X}_{1} , {X}_{2}
\\
\mbox{represent}~{N}~\mbox{o}&\mbox{r}~{P}.~\mbox{As~specified~in~the~polarity~model,~we~have}~P(\overline{\mathcal{X}}| w_i^j ) = 1 - P(\mathcal{X}|w_i^j).
\end{aligned}
\]
}
\caption{Inference rules with the polarity model.}
\label{fig:model_inference_rule}
\end{center}
\end{figure}

The glue rules ${X} \rightarrow {X}_{1} {X}_{2}$ combine two spans ($w_i^k$, $w_k^j$). The polarity value is calculated by their product, and normalized to 1.
\begin{equation}
\Phi :
		\begin{cases}
			P(\mathcal{X}|w_i^j) = \frac { P(\mathcal{X}|w_i^k) P(\mathcal{X}|w_k^j) }{ P(\mathcal{X}|w_i^k) P(\mathcal{X}|w_k^j)+ P(\overline{\mathcal{X}}|w_i^k) P(\overline{\mathcal{X}}|w_k^j) } \\
			P(\overline{\mathcal{X}}|w_i^j) = 1 - P(\mathcal{X}|w_i^j)
		\end{cases}
\end{equation}

For OOV text spans, the polarity model does not calculate the polarity values. When they are combined with in-vocabulary phrases by the auxiliary rules, the polarity values are determined by the text span with polarity and the OOV text span is ignored. More specifically,
\begin{equation}
\Phi :
		\begin{cases}
			P(\mathcal{X}| w_i^j ) = {P}(\mathcal{X}| w_i^k ) \\
			P(\overline{\mathcal{X}}| w_i^j ) = {P}(\overline{\mathcal{X}}| w_i^k )
		\end{cases}
\end{equation}

%

The combination rules are more complicated than other types of rules. In this article, we model the polarity probability calculation as the logistic regression. The logistic regression can be regarded as putting linear combination of the sub-spans' polarity probabilities into a logistic function (or sigmoid function). We will show that the negation, intensification, and contrast can be well modeled by the regression based method. It is formally shown as,
\begin{equation}
\label{eq:logistic_model_func}
\begin{aligned}
P(\mathcal{X}|w_i^j)
&= h\left( \mathbf{\theta}_0 + \sum _{ k=1 }^{ K }{ \mathbf{\theta}_k P(\mathcal{X}_k | w_{i_k}^{j_k})} \right) \\
&= \frac{1}{1+\exp \left\{ - \left( \mathbf{\theta}_0 + \sum _{ k=1 }^{ K }{ \mathbf{\theta}_k P(\mathcal{X}_k | w_{i_k}^{j_k})} \right) \right\}}
\end{aligned}
\end{equation}
where $h(x)=\frac{1}{1+\exp{\{-x\}}}$ is the logistic function, $K$ is the number of non-terminals in a rule, and $\mathbf{\theta}_0 , \dots , \mathbf{\theta}_K$ are the parameters that are learned from data. 
As a concrete example, if the span $w_i^j$ can match ${N} \rightarrow \mbox{not}~P$ and $P \overset{*}{\Rightarrow} w_{i+1}^j$, the inference rule with the polarity model is defined as,
\begin{equation}
\frac { {N} \rightarrow \mbox{not}~P  \quad [i+1,{P},j] \Phi_1 }
{ [i ,{N}, j]  
\begin{cases}
	P(\mathcal{N}|w_i^j) = h( \mathbf{\theta}_0 + \mathbf{\theta}_1 P(\mathcal{P}|w_{i+1}^j) ) \\
	P(\mathcal{P}|w_i^j) = 1 - P( \mathcal{N} | w_i^j )
\end{cases} }
\end{equation}
where polarity probability is calculated by $P(\mathcal{N}|w_i^j)= h (\mathbf{\theta}_0 + \mathbf{\theta}_1 P(\mathcal{P}|w_{i+1}^j))$.

To tackle negation, \textbf{switch negation}~\cite{sauri2008factuality,Choi:2008:compostional} simply reverses the sentiment polarity and corresponding sentiment strength. However, consider ``\textit{not great}'' and ``\textit{not good}'', flipping polarity directly makes ``\textit{not good}'' more positive than ``\textit{not great}'', which is unreasonable. Another potential problem of switch negation is that negative polarity items interact with intensifiers in undesirable ways~\cite{KennedyCI2006}. For example, ``\textit{not very good}'' turns out to be even more negative than ``\textit{not good}'', given the fact that ``\textit{very good}'' is more positive than ``\textit{good}''.
Therefore, \namecite{cl:lexicon} argue that \textbf{shift negation} is a better way to handle polarity negation. Instead of reversing polarity strength, shift negation shifts it toward the opposite polarity by a fixed amount. This method can partially avoid the aforementioned two problems. However, they set the parameters manually which might not be reliable and extensible enough to a new dataset.
Employing the regression model, switch negation is captured by the negative scale item $\mathbf{\theta}_k$ ($k>0$), and shift negation is expressed by the shift item $\mathbf{\theta}_0$.

The intensifiers are adjectives or adverbs which strengthen (amplifier) or decrease (downtoner) the semantic intensity of its neighboring item~\cite{quirk1985comprehensive}. For example, ``\textit{extremely good}'' should obtain higher strength of positive polarity than ``\textit{good}'', because it is modified by the amplifier (``\textit{extremely}'').
\namecite{polanyi2006contextual,KennedyCI2006} handle intensifiers by polarity addition and subtraction. This method, termed \textbf{fixed intensification}, increases a fixed amount of polarity for amplifiers and decreases for downtoners.
\namecite{cl:lexicon} propose a method, called \textbf{percentage intensification}, to associate each intensification word with a percentage scale, which is larger than one for amplifiers, and less than one for downtoners.
The regression model can capture these two methods to handle the intensification. The shift item $\mathbf{\theta}_0$ represents the polarity addition and subtraction directly, and the scale item $\mathbf{\theta}_k$ ($k>0$) can scale the polarity by a percentage.

\begin{table}[tb]
\caption{The check mark means the parameter of the polarity model can capture the corresponding intensification type and negation type. Shift item $\theta_0$ can handle shift negation and fixed intensification, and scale item $\theta_1$ can model switch negation and percentage intensification. 
}
\begin{tabular*}{\textwidth}{l c c l c c}
\hline
\multirow{2}{*}{Parameter} & \multicolumn{2}{c}{\tabincell{c}{Negation Type \\ $P(\mathcal{X}| w_{i}^{j})=h ( \mathbf{\theta}_0 + \mathbf{\theta}_1 P(\overline{\mathcal{X}}| w_{i_1}^{j_1}) )$}}  & &
\multicolumn{2}{c}{\tabincell{c}{Intensification Type \\ $P(\mathcal{X}| w_{i}^{j})= h ( \mathbf{\theta}_0 + \mathbf{\theta}_1 P(\mathcal{X}| w_{i_1}^{j_1}) )$}}
\\
\cline{2-3}
\cline{5-6}
 & Shift & Switch & & Percentage & Fixed \\ \hline
$\mathbf{\theta}_0$ (Shift item) & \checkmark &   & &   & \checkmark \\
$\mathbf{\theta}_1$ (Scale item) &   & \checkmark & & \checkmark &   \\
\hline
\end{tabular*}
\label{table:computation_model:negation_intensification}
\end{table}

Table~\ref{table:computation_model:negation_intensification} illustrates how the regression based polarity model represents different negation and intensification methods.
For a specific rule, the parameters and the compositional method are automatically learned from data (Section~\ref{sec:learn_computation_model}) instead of setting them manually as in previous work~\cite{cl:lexicon}.
In a similar way, this method can handle the contrast. For example, the inference rule for ${N} \rightarrow {P}~but~N$ is:
\begin{equation}
\label{eq:rule_pbutn}
\frac { ( {N} \rightarrow {P}~\mbox{but}~N )  \quad  [ {i}_{1} ,{P},{j}_{1}] \Phi_1 \quad [ {i}_{2} ,{N}, {j}_{2}] \Phi_2 }
{ [i,{N},j]
\begin{cases}
	P(\mathcal{N}|w_i^j)= h( \mathbf{\theta}_0 +{ \mathbf{\theta}_1 P(\mathcal{P} | w_{i_1}^{j_1})} + { \mathbf{\theta}_2 P(\mathcal{N} | w_{i_2}^{j_2})} ) \\
	P(\mathcal{P} |w_i^j) = 1 - P(\mathcal{N}|w_i^j)
\end{cases}  }
\end{equation}
where the polarity probability of the rule ${N} \rightarrow {P}~but~N$ is computed by $P(\mathcal{N}|w_i^j)=h (\mathbf{\theta}_0 +{ \mathbf{\theta}_1 P(\mathcal{P} | w_{i_1}^{j_1})} + { \mathbf{\theta}_2 P(\mathcal{N} | w_{i_2}^{j_2})} )$. It can express the contrast relation by specific parameters $\mathbf{\theta}_0$, $\mathbf{\theta}_1$, and $\mathbf{\theta}_2$.

\begin{figure}[t]
\begin{center}
\includegraphics[width=0.46\textwidth]{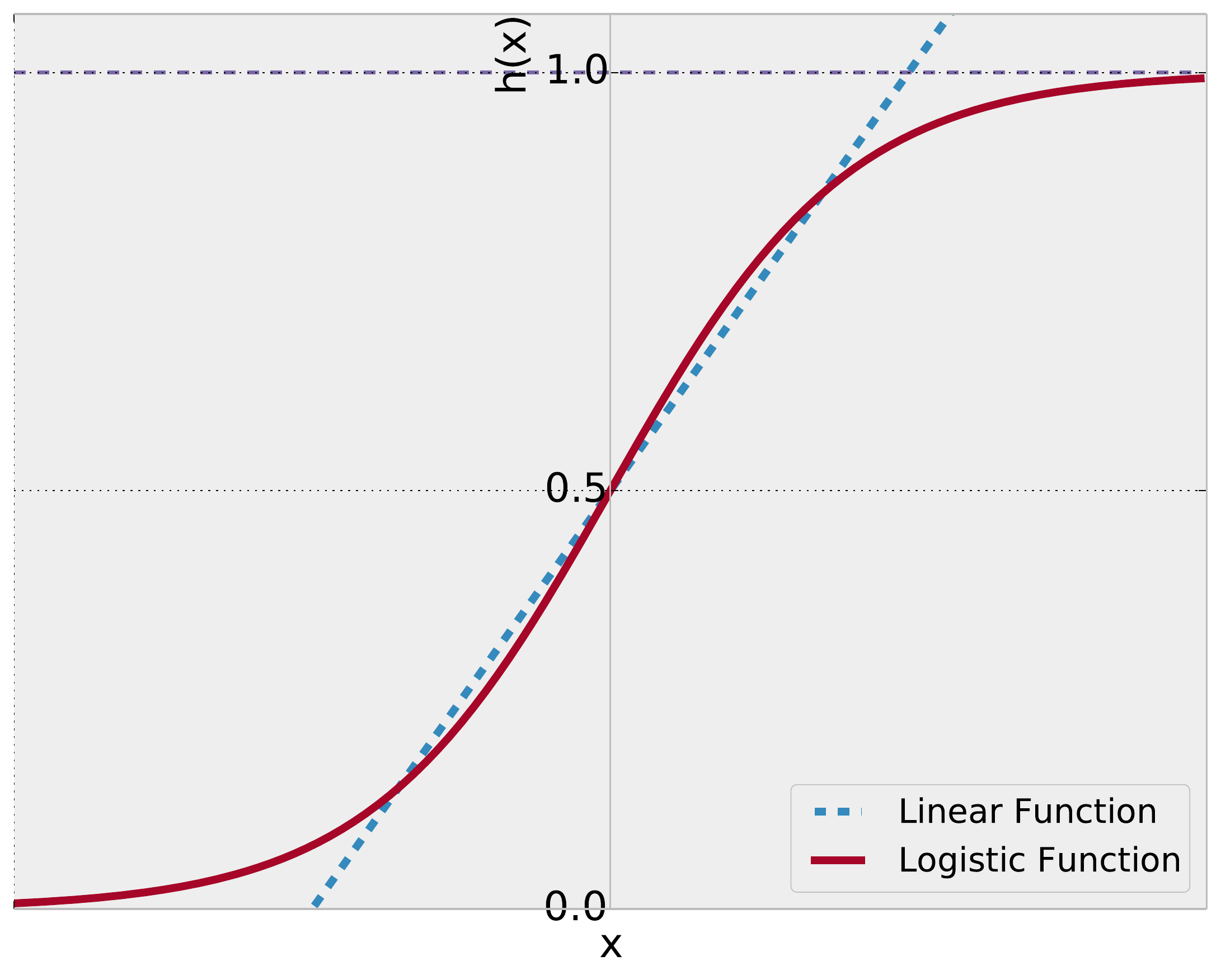}
\caption{Logistic function $h(x)=\frac{1}{1+\exp\{ -x \}}$ truncates polarity values to $(0,1)$ smoothly. The computed values are used as polarity probabilities.}
\label{fig:logistic_function}
\end{center}
\end{figure}

It should be noted that a linear regression model could turn out to be problematic, as it may produce unreasonable results. For example, if we do not add any constraint, we may get $P(\mathcal{N}|w_i^j)=-0.6+P(\mathcal{P}|w_{i+1}^{j})$. When $P(\mathcal{P}|w_{i+1}^{j})=0.55$, we will get $P(\mathcal{N}|w_i^j)=-0.6+0.55=-0.05$. It conflicts with the definition that the polarity probability ranges from zero to one. Figure~\ref{fig:logistic_function} intuitively shows that the logistic function truncates polarity values to $(0,1)$ smoothly.

\subsection{Constraints} \label{sec:constraints}
We incorporate additional constraints into the parsing model. They are used as pruning conditions in the derivation process not only to improve efficiency but also to force the derivation towards the correct direction.
We expand the inference rules in Section~\ref{sec:polarity_model} as,
\begin{equation}
\frac { (r) \quad {H}_1 \Phi_1  \quad \dots \quad {H}_K \Phi_K }{ [i , X , j] \Phi } C
\end{equation}
where $C$ is a \textbf{side condition}. The constraints are interpreted in a Boolean manner. If the constraint $C$ is satisfied, the rule can be used, otherwise, it cannot. We define two constraints in the parsing model.

First, in the parsing process, the polarity label of text span $w_i^j$ obtained by the polarity model (Section~\ref{sec:polarity_model}) should be consistent with the non-terminal $X$ ($N$ or $P$) on the left hand side of the rule. To distinguish between the polarity labels and the non-terminals, we denote the corresponding polarity label of non-terminal $X$ as $\mathcal{X}$. Following this notation, we describe the first constraint as,
\begin{equation}\label{eq:constraint:first}
C_1 : P(\mathcal{X} | w_i^j ) > P(\overline{\mathcal{X}} | w_i^j )
\end{equation}
where $\overline{\mathcal{X}}$ is the opposite polarity of $\mathcal{X}$. For instance, if rule $P \rightarrow not \ N$ matches the text span $w_i^j$, the polarity calculated by the polarity model should be consistent with $P$, i.e., the polarity obtained by the polarity model should be positive ($\mathcal{P}$).

Second, when we apply the combination rules, the polarity strength of sub-spans needs to exceed a predefined threshold $\tau$ ($\ge 0.5$). Specifically, for combination rules ${X}\rightarrow {w}^{i_1}_{i} {X}_{1} {w}^{i_2}_{j_1} {X}_{2} {w}^{j}_{j_2}$ and ${X} \rightarrow {w}^{i_1}_{i} {X}_{1} {w}^{j}_{j_1}$, we define the second constraint as,
\begin{equation}
C_2 : P( \mathcal{X}_k | w_{i_k}^{j_k} ) > \tau , k=1, \dots , K
\end{equation}
where $K$ is the number of sub-spans in the rule, and $\mathcal{X}_k$ is the corresponding polarity label of non-terminal $X_k$ in the right hand side.
If $P( \mathcal{X}_k | w_{i_k}^{j_k} )$ is not larger than threshold $\tau$, we regard the polarity of phrase $w_{i_k}^{j_k}$ as neutral. For instance, we do not want to use the combination rule $P \rightarrow \mbox{a~lot~of}~P$ or $N \rightarrow \mbox{a~lot~of}~N$ for the phrase ``\textit{a lot of people}''. This constraint avoids improperly using the combination rules for neutral phrases.
Notably, when $\tau$ is set as 0.5, this constraint is the same as the first one in~\eqref{eq:constraint:first}.

As shown in Figure~\ref{fig:model_inference_rule_constraint}, we add these two constraints to the inference rules. The OOV rules do not have any constraints, and the constraint $C_1$ is applied for all the other rules. The constraint $C_2$ is only applied for the combination rules.

\begin{figure}[t]
\begin{center}
\[
\begin{aligned}
&\frac { ({X} \rightarrow w^j_i ) }{ [i , {X} , j]
		P(\mathcal{X}|w_i^j) = \tilde{P}(\mathcal{X}|w_i^j)
	} C_1
\\
&\frac { ({X} \rightarrow {w}^{i_1}_{i} {X}_{1} {w}^{j}_{j_1} ) \quad  [ {i}_{1} ,{X}_{1}  , {j}_{1} ] \Phi_1 }{ [i,{X},j]
		P(\mathcal{X}|w_i^j) = h( \mathbf{\theta}_0 + \mathbf{\theta}_1 P(\mathcal{X}_{1} | w_{i_1}^{j_1}) )  } C_1 \wedge C_2
\\
&\frac {( {X}\rightarrow {w}^{i_1}_{i} {X}_{1} {w}^{i_2}_{j_1} {X}_{2} {w}^{j}_{j_2} )  \quad  [ {i}_{1} , {X}_{1} ,{j}_{1}] \Phi_1  \quad [ {i}_{2} ,{X}_{2} , {j}_{2}] \Phi_2 }{ [i,{X},j]
		P(\mathcal{X}|w_i^j) = h( \mathbf{\theta}_0 + { \mathbf{\theta}_1 P(\mathcal{X}_1 | w_{i_1}^{j_1})} + { \mathbf{\theta}_2 P(\mathcal{X}_2 | w_{i_2}^{j_2})} ) } C_1 \wedge C_2
\\
&\frac { ({X}\rightarrow {X}_{1} {X}_{2} ) \quad [i, {X}_{1} ,k] \Phi_1  \quad [k, {X}_{2} ,j] \Phi_2 }{ [i,{X},j]
		P(\mathcal{X}|w_i^j) = \frac { P(\mathcal{X}|w_i^k) P(\mathcal{X}|w_k^j) }{ P(\mathcal{X}|w_i^k) P(\mathcal{X}|w_k^j)+ P(\overline{\mathcal{X}}|w_i^k) P(\overline{\mathcal{X}}|w_k^j) } } C_1
\\
&\frac { (\mathcal{E} \rightarrow w^j_i ) }{ [i , \mathcal{E} , j]
	\circ
} \circ
\\
&\frac { ({X}\rightarrow \mathcal{E} {X}_{1} ) \quad [i, \mathcal{E} ,k] \circ  \quad [k, {X}_{1} ,j] \Phi_1 }{ [i,{X},j]
		P(\mathcal{X}|w_i^j) =  P(\mathcal{X}|w_k^j) } C_1
\\
&\frac {( {X}\rightarrow {X}_{1} \mathcal{E} ) \quad [i, {X}_{1} ,k] \Phi_1  \quad [k, \mathcal{E} ,j] \circ }{ [i,{X},j]
		P(\mathcal{X}|w_i^j) = P(\mathcal{X}|w_i^k) } C_1
\\
\mbox{where}~h(x) =& \frac{1}{1+\exp\{ -x \}}~\mbox{is~a~logistic~function,} \circ \mbox{represents~the~absence, and}~{X} , {X}_{1} , {X}_{2}
\\
\mbox{represent}~{N}~\mbox{o}&\mbox{r}~{P}.~\mbox{As~specified~in~the~polarity~model,~we~have}~P(\overline{\mathcal{X}}| w_i^j ) = 1 - P(\mathcal{X}|w_i^j).
\end{aligned}
\]
\caption{Inference rules with the polarity model and constraints.}
\label{fig:model_inference_rule_constraint}
\end{center}
\end{figure}

\subsection{Decoding Algorithm}
In this section, we summarize the decoding algorithm in Algorithm~\ref{alg:decoding}. For a sentence $s$, the CYK algorithm and dynamic programming are employed to obtain the sentiment tree with the highest score.
To be specific, the modified CYK parsing model parses the input sentence to sentiment trees in a bottom-up way, i.e., from short to long text spans. For every text span $w_i^j$, we match the rules in the sentiment grammar (Section~\ref{sec:sentiment_grammar}) to generate the candidate set. Their polarity values are calculated using the polarity model described in Section~\ref{sec:polarity_model}. We also employ the constraints described in Section~\ref{sec:constraints} to prune search paths. The constraints improve the efficiency of the parsing algorithm and make derivations that meet our intuitions.

The features in the ranking model (Section~\ref{sec:feature}) decompose along the structure of the sentiment tree. So the dynamic programming technique can be used to compute the derivation tree with the highest ranking score. For a span, the scores of its subspans are used to calculate the local scores of its derivations.
For example, the score of the derivation $ \frac { ( r ) ~~ [ {i}_{1} ,{P},{j}_{1}] ~~ [ {i}_{2} ,{N}, {j}_{2}] } { [i,{X},j] } $ is $score[{i_1},{j_1},{P}] + score[{i_2},{j_2},{N}] + score^{r}$, where $score[{i},{j},{X}]$ is the highest score of text span $w_i^j$ which is derived to the non-terminal $X$, and $score^{r}$ is the score of applying the rule $r$.
As described in Section~\ref{sec:ranking_model}, the score of using rule $r$ is $score^{r} = {\mathbf{\phi}( w_i^j , r )}^{\mathsf{T}} \mathbf{\psi}$, where ${\mathbf{\phi}( w_i^j , r )}$ is the feature vector of using the rule $r$ for the span $w_i^j$, and $\mathbf{\psi}$ is the weight vector of the ranking model.
The $k$ highest score trees satisfying the constraints are stored in $score[,,]$ for decoding the longer text spans.
After finishing the CYK parsing, $\argmax_{\mathcal{X} \in \{\mathcal{N},\mathcal{P}\}}{score[0,n,X]}$ is regarded as the polarity label of input sentence.
The time complexity is the same as the standard CYK's.


\begin{algorithm*}[t]
\caption{Decoding Algorithm
\label{alg:decoding}}
\begin{algorithmic}[1]
\Require $w_0^n$: Sentence 
\Ensure Polarity of the input sentence

\Let{$score[,,]$}{$\{ \}$}
\For{$l \gets 1 \dots n$} \Comment{Modified CYK algorithm}
	\ForAll{$i,j ~~ s.t. ~~ j-i=l$}
		\ForAll{inferable rule $ \frac { ( r ) ~~ H_1 \dots H_K } { [i,{X},j] } $ for $w_i^j$}
			\Let{$\Phi$}{calculate polarity value for $r$} \Comment{Polarity model}
			\If{constraints are satisfied} \Comment{Constraint}
				\Let{$sc$}{compute score for this derivation by ranking model} \Comment{Ranking model}
				\If{$sc > score[i,j,X]$}
					\Let{$score[i,j,X]$}{$sc$}
				\EndIf
			\EndIf
		\EndFor
	\EndFor
\EndFor
\State \Return{$\argmax_{\mathcal{X} \in \{\mathcal{N},\mathcal{P}\}}{score[0,X,n]}$}
\end{algorithmic}
\end{algorithm*}



\section{Model Learning} \label{sec:model_learning}
We have described the statistical sentiment parsing framework in the above section. We present the model learning process in this section.
The learning process consists of two steps.
First, the sentiment grammar and the polarity model are learned from data. In other words, the rules and the parameters used to compute polarity values are learned. These basic sentiment building blocks are then used to build the parse trees.
Second, we estimate the parameters of the ranking model using the sentence and polarity label pairs. In this stage, we concentrate on learning how to score the parse trees based on the learned sentiment grammar and polarity model.

Section~\ref{sec:ranking_model_learning} shows the features and the parameter estimation algorithm used in the ranking model. Section~\ref{sec:grammar_learning} illustrates how to learn the sentiment grammar and the polarity model.

\subsection{Ranking Model Training} \label{sec:ranking_model_learning}
As shown in Section~\ref{sec:ranking_model}, we develop the ranking model upon the log-linear model. In the following sub-sections, we first present the features used to rank sentiment tree candidates. Then, we describe the objective function used in the optimization algorithm. Finally, we introduce the algorithm for parameter estimation using the gradient-based method.

\subsubsection{Features} \label{sec:feature}
\begin{figure}[tb]
\begin{center}
\subfloat[\textsc{CombHit} and \textsc{CombRule}\label{fig:features_ranking_model:a}]{%
	\includegraphics[width=0.47\textwidth]{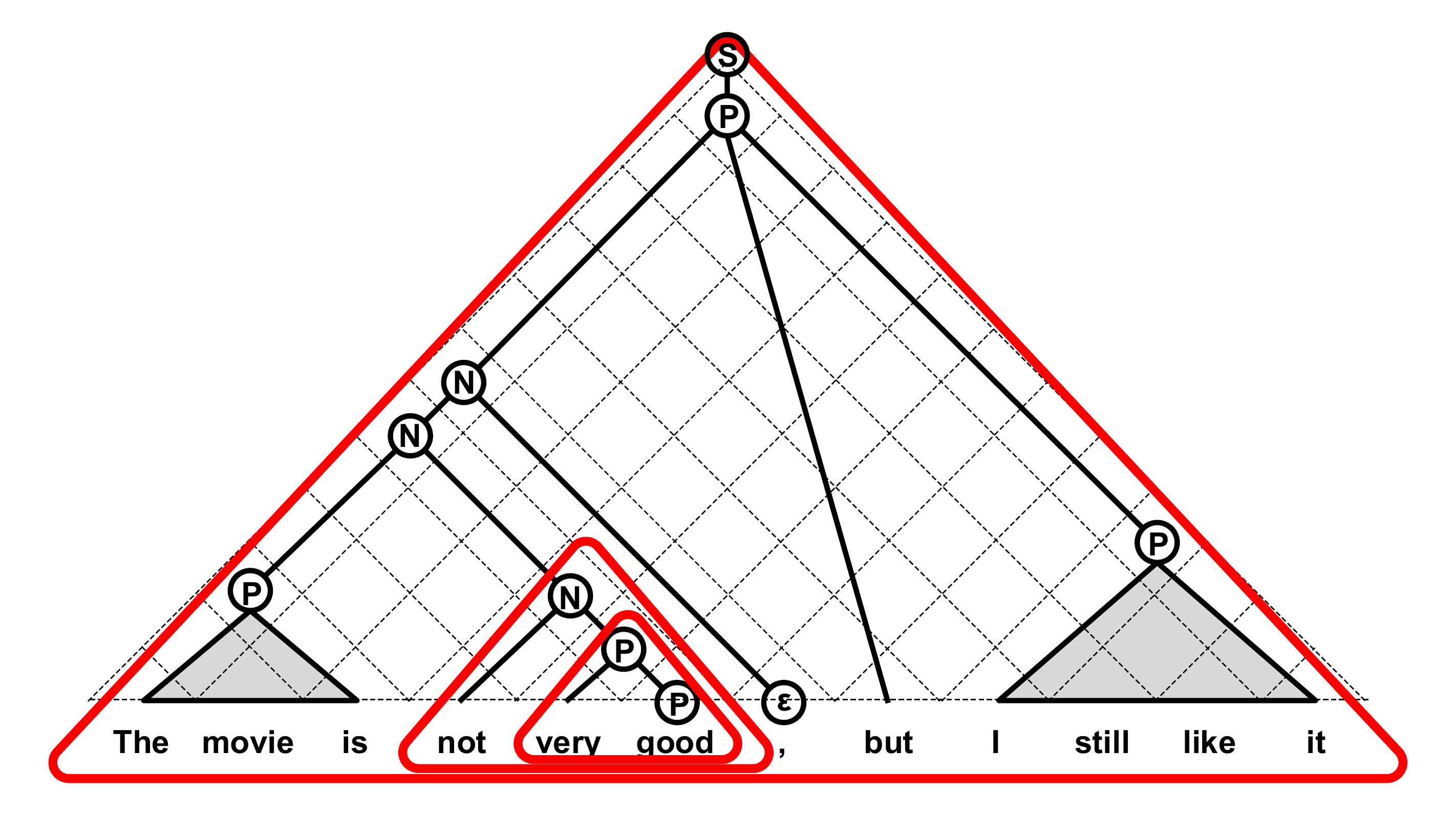}
}
\hfill
\subfloat[\textsc{DictHit} and \textsc{DictRule}\label{fig:features_ranking_model:b}]{%
	\includegraphics[width=0.47\textwidth]{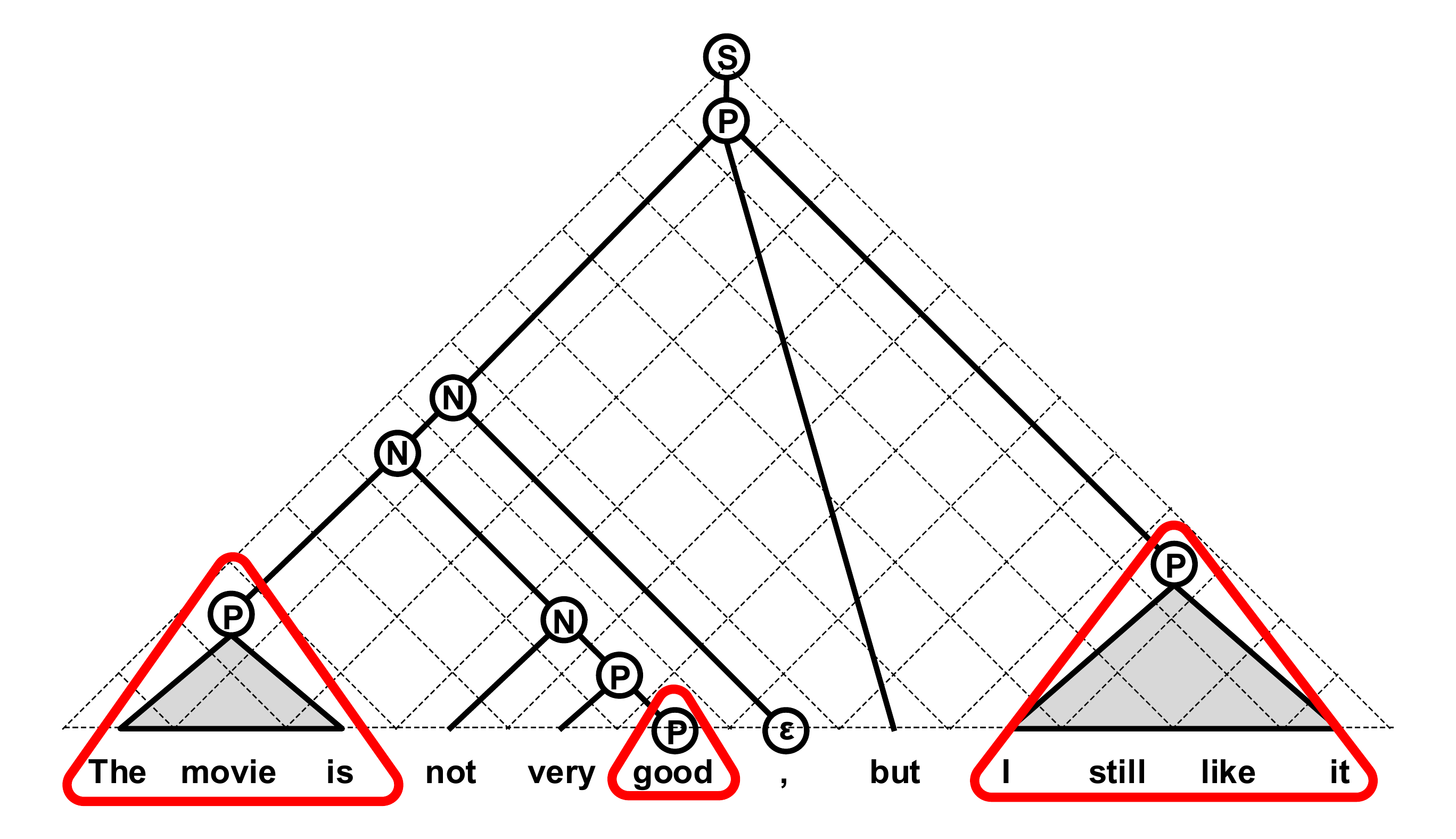}
}
\hfill
\begin{tabular}{l l l}
\\
\textbf{Feature Template}	& \textbf{Feature}	& \textbf{Feature Value} \\
Number of combination rules	& \textsc{CombHit}  & 3 \\
Combination rule			& $\textsc{CombRule}[{P} \rightarrow \mbox{very}~P]$  & 1 \\
~							& $\textsc{CombRule}[{N} \rightarrow \mbox{not}~P]$  & 1 \\
~							& $\textsc{CombRule}[{P} \rightarrow {N} ~\mbox{but}~ P]$  & 1 \\
Number of dictionary rules	& \textsc{DictHit}  & 3 \\
Dictionary rule				& $\textsc{DictRule}[{P} \rightarrow \mbox{the~movie~is}]$  & 1 \\
~							& $\textsc{DictRule}[{P} \rightarrow \mbox{good}]$  & 1 \\
~							& $\textsc{DictRule}[{P} \rightarrow \mbox{i~ still~ like~ it}]$  & 1 \\
\end{tabular}
\caption{Feature templates used in the ranking model. The red triangles denote the features for the example.}
\label{fig:features_ranking_model}
\end{center}
\end{figure}

We extract a feature vector $\mathbf{\phi}(s,t) \in \mathcal{R}^{d}$ for each parse tree $t$ of sentence $s$. The feature vector is used in the log-linear model.
In Figure~\ref{fig:features_ranking_model}, we present the features extracted for the sentence ``\emph{The movie is not very good, but i still like it}''.
The features are organized into feature templates. Each of them contains a set of features. These feature templates are shown as follows:

\begin{itemize}
\item \textsc{CombHit}: This feature is the total number of combination rules used in $t$.
\item \textsc{CombRule}: It contains features $\{ \textsc{CombRule}[r] : r~\mbox{is}~\mbox{a}~\mbox{combination}~\mbox{rule} \}$, each of which fires on the combination rule $r$ appearing in $t$.
\item \textsc{DictHit}: This feature is the total number of dictionary rules used in $t$.
\item \textsc{DictRule}: It contains features $\{ \textsc{DictRule}[r] : r~\mbox{is}~\mbox{a}~\mbox{dictionary}~\mbox{rule} \}$, each of which fires on the dictionary rule $r$ appearing in $t$.
\end{itemize}

These features are generic local patterns which capture the properties of the sentiment tree.
Another intuitive lexical feature template is [combination rule + word]. For instance, ${P} \rightarrow \mbox{very}~P(\mbox{good})$ is a feature which lexicalizes the non-terminal $P$ to $\mbox{good}$. However, if this feature is fired frequently, the phrase ``\textit{very good}'' would be learned as a dictionary rule and can be used in the decoding process. So we do not employ this feature template in order to reduce the feature size.
It should be noted that these features decompose along structures of sentiment trees, enabling us to use dynamic programming in the CYK algorithm.

\subsubsection{Objective Function}
We design the ranking model upon the log-linear model to score candidate sentiment trees.
In the training data $\mathcal{D}$, we only have the input sentence $s$ and its polarity label $\mathcal{L}_{s}$. The forms of sentiment parse trees, which can obtain the correct sentiment polarity, are unobserved. So we work with the marginal log-likelihood of obtaining the correct polarity label $\mathcal{L}_{s}$, 
\begin{equation}
\begin{aligned}
\log {p(\mathcal{L}_{s}|s;T,\mathbf{\psi})} &= \log {p(t \in T^{\mathcal{L}_{s}} {(s)} |s;T,\mathbf{\psi})} \\
&=\mathbf{A}(\mathbf{\psi};s,T^{\mathcal{L}_{s}})-\mathbf{A}(\mathbf{\psi};s,T)
\end{aligned}
\end{equation}
where $T^{\mathcal{L}_{s}}$ is the set of candidate trees whose prediction labels are $\mathcal{L}_{s}$, and $\mathbf{A}(\mathbf{\psi};s,T)$ (Equation~\eqref{eq:log_linear_model_A}) is the log-partition function with respect to $T(s)$.

Based on the marginal log-likelihood function, the objective function $\mathcal{O}(\mathbf{\psi},T)$ consists of two terms. The first term is the sum of marginal log-likelihood over training instances which can obtain the correct polarity labels. The second term is a $L^2$-norm regularization term on the parameters $\mathbf{\psi}$. Formally,
\begin{equation}
\label{eq:obj_func}
\begin{aligned}
\mathcal{O}(\mathbf{\psi},T) &= \sum _{\substack{(s,\mathcal{L}_{s}) \in \mathcal{D} \\ T^{\mathcal{L}_{s}} {(s)} \neq \emptyset}}{\log {p(\mathcal{L}_{s}|s;T,\mathbf{\psi})}} - \frac { \lambda  }{ 2 } {\left\| \mathbf{\psi} \right\|}_{2}^{2} \\
\end{aligned}
\end{equation}

To learn the parameters $\mathbf{\psi}$, we employ a gradient-based optimization method to maximize the objective function $\mathcal{O}(\mathbf{\psi},T)$. According to \namecite{obj_grad}, the derivative of the log-partition function is the expected feature vector:
\begin{equation}
\label{eq:obj_func_grad}
\frac { \partial {\mathcal{O}(\mathbf{\psi},T)} }{ \partial {\mathbf{\psi}} } = \sum _{\substack{(s,\mathcal{L}_{s}) \in \mathcal{D} \\ T^{\mathcal{L}_{s}} {(s)} \neq \emptyset}} {({E}_{p(t|s;T^{\mathcal{L}_{s}},\mathbf{\psi})}{[ \mathbf{\phi}(s,t) ]} - {E}_{p(t|s;T,\mathbf{\psi})}{[ \mathbf{\phi}(s,t) ]})} - \lambda \mathbf{\psi}
\end{equation}
where $E_{p(x)}{[f(x)]}=\sum_{x}{p(x)f(x)}$ for discrete $x$.

\subsubsection{Parameter Estimation}
The objective function $\mathcal{O}(\mathbf{\psi},T)$ is not concave (nor convex), hence the optimization potentially results in a local optimum.
Stochastic Gradient Descent (SGD;~\citewop{Robbins&Monro:1951}) is a widely used optimization method. The SGD algorithm picks up a training instance randomly, and updates the parameter vector $\mathbf{\psi}$ according to
\begin{equation}
\label{eq:basic_sgd}
{\mathbf{\psi}_j}^{(t+1)} = {\mathbf{\psi}_j}^{(t)} + \alpha \left( \frac { \partial {\mathcal{O}(\mathbf{\psi})} }{ \partial {\mathbf{\psi}_j} } |_{\mathbf{\psi} = {\mathbf{\psi}}^{(t)} } \right)
\end{equation}
where $\alpha$ is the learning rate, and $\frac { \partial {\mathcal{O}(\mathbf{\psi})} }{ \partial {\mathbf{\psi}_j} }$ is the gradient of the objective function with respect to parameter $\mathbf{\psi}_j$. The SGD is sensitive to $\alpha$, and the learning rate is the same for all dimensions.
As described in Section~\ref{sec:feature}, we mix sparse features together with dense features. We want the learning rate to be different for each dimension.
We employ AdaGrad~\cite{Duchi:2011:ASM:1953048.2021068} to update the parameters, which sets an adaptive per-feature learning rate. The AdaGrad algorithm tends to use smaller update steps when we meet a feature many times. In order to compute efficiently, a diagonal approximation version of AdaGrad is used. The update rule is
\begin{equation}
\label{eq:adagrad}
\begin{aligned}
{\mathbf{\psi}_j}^{(t+1)} &= {\mathbf{\psi}_j}^{(t)} + \alpha \frac { 1 }{ \sqrt { { G }_{ j }^{ (t+1) } } } \left( \frac { \partial {\mathcal{O}(\mathbf{\psi})} }{ \partial {\mathbf{\psi}_j} } |_{\mathbf{\psi} = {\mathbf{\psi}}^{(t)} } \right) \\
{ { G }_{ j }^{ (t+1) } } &= { { G }_{ j }^{ (t) } } + { \left( \frac { \partial { \mathcal{ O }({ \psi  }) } }{ \partial { { \psi  }_{ j } } } |_{ { \psi  }={ { \psi  } }^{ (t) } } \right)  }^{ 2 }
\end{aligned}
\end{equation}
where we introduce an adaptive term ${ G }_{ j }^{ (t) }$. ${ G }_{ j }^{ (t) }$ becomes larger along with updating, and decreases the update step for dimension $j$. Compared to SGD, the only cost is to store and update ${ G }_{ j }^{ (t) }$ for each parameter.

To train the model, we use the method proposed by~\namecite{percyliang}.
With the candidate parse trees and objective function, the parameters $\mathbf{\psi}$ are updated to make the parsing model favor correct trees and give them a higher score.
Because there are many parse trees for a sentence, we need to calculate Equation~\eqref{eq:obj_func_grad} efficiently.
As indicated in Section~\ref{sec:feature}, the features decompose along the structure of sentiment tree. So dynamic programming can be employed to compute ${E}_{p(t|s;T,\mathbf{\psi})}{[ \mathbf{\phi}(s,t) ]}$ of~\eqref{eq:obj_func_grad}.
However, the first expectation term ${E}_{p(t|s;T^{\mathcal{L}_{s}},\mathbf{\psi})}{[ \mathbf{\phi}(s,t) ]}$ sums over the candidates which obtain the correct polarity labels. As this constraint does not decompose along the tree structure, there is no efficient dynamic program for this.
Instead of searching all the parse trees spanning $s$, we use beam search to approximate this expectation. Beam search is a best-first search algorithm which explores at most $K$ paths ($K$ is the beam size). It keeps the local optimums to reduce the huge search space.
Specifically, the beam search algorithm generates the K-best trees with the highest score $\mathbf{\phi}(s,t)^{\mathsf{T}} \mathbf{\psi}$ for each span. These local optimums are used recursively in the CYK process. The K-best trees for the whole span are regarded as the candidate set $\tilde{T}$. Then $\tilde{T}$ and $\tilde{T}^{\mathcal{L}_{s}}$ are used to approximate Equation~\eqref{eq:obj_func_grad} as in~\cite{percyliang}.

The intuition behind this parameter estimation algorithm lies in: (1) if we have better parameters, we can obtain better candidate trees; (2) with better candidate trees, we can learn better parameters. Thus the optimization problem is solved in an iterative manner. We initialize the parameters as zeros. This leads to a random search and generates random candidate trees. With the initial candidates, the two steps in Algorithm~\ref{alg:ranking_model_learning} lead the parameters $\mathbf{\psi}$ towards the direction achieving better performance.

\begin{algorithm}[t]
\caption{Ranking Model Learning Algorithm
\label{alg:ranking_model_learning}}
\begin{algorithmic}[1]
\Require $\mathcal{D}$: Training data $\{( s , \mathcal{L}_s )\}$, $S$: Maximum number of iteration
\Ensure $\mathbf{\psi}$: Parameters of the ranking model

\Let {$\mathbf{\psi}^{(0)}$} {${(0,0, \dots ,0)}^{\mathsf{T}}$}
\Repeat
	\Let{$( s , \mathcal{L}_s )$}{randomly select a training instance in $\mathcal{D}$}
	\Let {${\tilde {T}^{(t)}}$} {${\beamsearch}{(s , \mathbf{\psi}^{(t)})}$} \Comment{Beam search to generate K-best candidates}
	\Let {${ G }_{ j }^{ (t+1) }$} {${ { G }_{ j }^{ (t) } } + { \left( \frac { \partial { \mathcal{ O }({ \psi  } , {\tilde { T}^{(t)}}) } }{ \partial { { \psi  }_{ j } } } |_{ { \psi  }={ { \psi  } }^{ (t) } } \right)  }^{ 2 }$}
	\Let {$\mathbf{\psi}_{j}^{(t+1)}$}{${\mathbf{\psi}_j}^{(t)} + \alpha \frac { 1 }{ \sqrt { { G }_{ j }^{ (t+1) } } } \left( \frac { \partial {\mathcal{O}(\mathbf{\psi} , {\tilde { T}^{(t)}})} }{ \partial {\mathbf{\psi}_j} } |_{\mathbf{\psi} = {\mathbf{\psi}}^{(t)} } \right)$} \Comment{Update parameters using AdaGrad}
	\Let{$t$}{$t+1$}
\Until {$t > S$}
\State \Return {$\mathbf{\psi}^{(T)}$}
\end{algorithmic}
\end{algorithm}

\subsection{Sentiment Grammar Learning} \label{sec:grammar_learning}
In this section, we present the automatic learning of the sentiment grammar as defined in Section~\ref{sec:sentiment_grammar}. We need to extract the dictionary rules and the combination rules from data. In traditional statistical parsing, grammar rules are induced from annotated parse trees (such as the Penn TreeBank), so ideally we need examples of sentiment structure trees, or sentences annotated with sentiment polarity for the whole sentence as well as those for constituents within sentences.
However, this is not practical, if not unfeasible, as the annotations will be inevitably time consuming and require laborious human effort. In this article, we show that it is possible to induce the sentiment grammar directly from examples of sentences annotated with sentiment polarity labels without using any syntactic annotations or polarity annotations of constituents within sentences.
The sentences annotated with sentiment polarity labels are relatively easy to obtain, and we use them as our input to learn dictionary rules and combination rules.

We first present the basic idea behind the algorithm we proposed. People are likely to express positive or negative opinions using very simple and straightforward sentiment expressions again and again in their reviews. Intuitively, we can mine dictionary rules from these massive review sentences by leveraging the redundancy characteristic. Furthermore, there are many complicated reviews which contains complex sentiment structures (e.g., negation, intensification, and contrast). If we already have dictionary rules on hand, we can use them to obtain basic sentiment information for the fragments within complicated reviews. We can then extract combination rules with the help of the dictionary rules and the sentiment polarity labels of complicated reviews. Because the simple and straightforward sentiment expressions are often coupled with complicated expressions, we need to conduct dictionary rule mining and the combination rule mining in an iterative way.


\subsubsection{Dictionary Rule Learning} \label{sec:dictionary_rule_learning}
The dictionary rules $\mathcal{G_D}$ are basic sentiment building blocks used in the parsing process. Each dictionary rule in $\mathcal{G_D}$ is in the form $X \rightarrow f$, where $f$ is a sentiment fragment. We use the polarity probabilities $P(\mathcal{N}|f)$ and $P(\mathcal{P}|f)$ in the polarity model.
To build $\mathcal{G_D}$, we regard all the frequent fragments whose occurrence frequencies are larger than ${\tau}_{f}$ and lengths range from 1 to 7 as the sentiment fragments. We further filter the phrases formed by stop words and punctuations, which are not used to express sentiment.

For a balanced dataset, the sentiment distribution of a candidate sentiment fragment $f$ is calculated by,
\begin{equation}
\label{eq:dictionary_rule_computation}
P(\mathcal{X}|f) = \frac {\#(f, \mathcal{X}) + 1}{\#(f, \mathcal{N}) + \#(f, \mathcal{P}) + 2}
\end{equation}
where $\mathcal{X} \in \{ \mathcal{N} , \mathcal{P} \}$, and $\#(f, \mathcal{X})$ denotes the number of reviews containing $f$ with $\mathcal{X}$ being the polarity.
It should be noted that Laplace smoothing is used in Equation~\eqref{eq:dictionary_rule_computation} to deal with the zero frequency problem.

We do not learn the polarity probabilities $P(\mathcal{N}|f)$ and $P(\mathcal{P}|f)$ by directly counting occurrence frequency.
For example, in the review sentence ``\textit{this movie is not good}'' (negative), the naive counting method increases the count $\#(good , \mathcal{N})$ in terms of the polarity of the whole sentence.
Moreover, because of the common collocation ``\textit{not as good as}'' (negative) in movie reviews, ``\textit{as good as}'' is also regarded as negative if we count the frequency directly.
The examples indicate why some polarity probabilities of phrases counting from data are different from our intuitions. 
These unreasonable polarity probabilities also make trouble for learning the polarity model.
Consequently, in order to estimate more reasonable probabilities, we need to take the compositionality into consideration when learning sentiment fragments.

\begin{algorithm}[t]
\caption{Dictionary Rule Learning
\label{alg:dictionary_rule_learning}}
\begin{algorithmic}[1]
\Require $\mathcal{D}$: Dataset, $\mathcal{G_C}$: Combination rules, ${\tau}_{f}$: Frequency threshold
\Ensure $\mathcal{G_D}$: Dictionary rules

\Function{MineDictionaryRules}{$\mathcal{D} , \mathcal{G_C}$}
	\Let {$\mathcal{{G_D}'}$} {$\{\}$}
	\For {$(s,\mathcal{L}_{s})$ in $\mathcal{D}$} \Comment{$s : {w}_{0} {w}_{1} \cdots {w}_{ |s|-1 }$, $\mathcal{L}_{s}$: Polarity~label~of~$s$}
		\ForAll{$i,j ~~ s.t. ~~ 0 \le i < j \le |s|$} \Comment{${w}_{i}^{j} : {w}_{i} {w}_{i+1} \cdots {w}_{j-1}$}
			\If {no negation rule in $\mathcal{G_C}$ covers $w_i^j$}
				\AddOne {$\#({w}_{i}^{j},\mathcal{L}_{s})$}
			\EndIf
			\State {add ${w}_{i}^{j}$ to $\mathcal{{G_D}'}$}
		\EndFor
	\EndFor
	\Let {$\mathcal{G_D}$} {$\{\}$}
	\For{$f$ in $\mathcal{{G_D}'}$}
		\If {$\#(f,\cdot) \ge {\tau}_{f}$}
			\State {compute $P(\mathcal{N}|f)$ and $P(\mathcal{P}|f)$ using Equation~\eqref{eq:dictionary_rule_computation}}
			\State {add dictionary rule $({L}_{f} \rightarrow f)$ to $\mathcal{G_D}$} \Comment{${L}_{f} = \argmax_{{X} \in \{{N},{P}\}}{P(\mathcal{X}|f)}$}
		\EndIf
	\EndFor
	\State \Return{$\mathcal{G_D}$}
\EndFunction
\end{algorithmic}
\end{algorithm}

Following the above motivation, we ignore the count $\#(f, \mathcal{X})$, if the sentiment fragment $f$ is \textbf{covered} by a negation rule $r$ which negates the polarity of $f$.
The word ``\textbf{cover}'' here means that $f$ is derived within a non-terminal of the negation rule $r$.
For instance, the negation rule $N \rightarrow \mbox{not}~P$ covers the sentiment fragment ``\textit{good}'' in the sentence ``\textit{this is not a good movie}'' (negative), i.e., the ``\textit{good}'' is derived from $P$ of this negation rule. So we ignore the occurrence for $\#(\mbox{good}, \mathcal{N})$ in this sentence.
It should be noted that we still increase the count for $\#(\mbox{not}~\mbox{good}, \mathcal{N})$, because there is no negation rule covering the fragment ``\textit{not good}''.

As shown in Algorithm~\ref{alg:dictionary_rule_learning}, we learn the dictionary rules and their polarity probabilities by counting the frequencies in negative and positive classes. Only the fragments whose occurrence numbers are larger than threshold $\tau_f$ are kept.
Moreover, we take the combination rules into consideration to acquire more reasonable $\mathcal{G_D}$.
Notably, a subsequence of a frequent fragment must also be frequent. This is similar to the key insight in the Apriori algorithm~\cite{apriori}. When we learn the dictionary rules, we can count the sentiment fragments from short to long, and prune the infrequent fragments in the early stages if any subsequence is not frequent. This pruning method accelerates the dictionary rule learning process and makes the procedure fit in memory.

\subsubsection{Combination Rule Learning} \label{sec:combination_rule_learning}
The combination rules $\mathcal{G_C}$ are generalizations for the dictionary rules. They are used to handle the compositionality and process unseen phrases.
The learning of combination rules is based on the learned dictionary rules and their polarity values.
The sentiment fragments are generalized to combination rules by replacing the subsequences of dictionary rules with their polarity labels.
For instance, as shown in Figure~\ref{fig:computation_model}, the fragments ``\textit{is not (good/as expected/funny/well done)}'' are all negative. After replacing the sub-spans ``\textit{good}'', ``\textit{as expected}'', ``\textit{funny}'', and ``\textit{well done}'' with their polarity label $P$, we can learn the negation rule $N \rightarrow \mbox{is~not}~P$.


We present the combination rule learning approach in Algorithm~\ref{alg:combination_rule_learning}.
Specifically, the first step is to generate combination rule candidates. For every sub-sequence ${w}_{i}^{j}$ of sentiment fragment $f$, we replace it with the corresponding non-terminal ${L}_{{w}_{i}^{j}}$ if $P(\mathcal{L}_{{w}_{i}^{j}} | {w}_{i}^{j})$ is larger than the threshold $\tau_{p}$, and we can get ${w}_{0}^{i} {L}_{{w}_{i}^{j}} {w}_{j}^{|f|}$.
Next, we compare the polarity $\mathcal{L}_{{w}_{i}^{j}}$ with $\mathcal{L}_{f}$. If $\mathcal{L}_{f} \neq \mathcal{L}_{{w}_{i}^{j}}$, we regard the rule ${L}_{f} \rightarrow {w}_{0}^{i} {L}_{{w}_{i}^{j}} {w}_{j}^{|f|}$ as a negation rule. Otherwise, we further compare their polarity values. If this rule makes the polarity value become larger (or smaller), it will be treated as a strengthen (or weaken) rule.
To obtain the contrast rules, we replace two sub-sequences with their polarity labels in a similar way. If the polarities of these two sub-sequences are different, we categorize this rule to the contrast type. Notably, these two non-terminals can not be next to each other.
After the above steps, we get the rule candidate set $\mathcal{{G_C}'}$ and the occurrence number of each rule. We then filter the rule candidates whose occurrence frequencies are too small, and assign the rule types (negation, strengthen, weaken, and contrast) according to their occurrence numbers.

\begin{algorithm}[t]
\caption{Combination Rule Learning
\label{alg:combination_rule_learning}}
\begin{algorithmic}[1]
\Require $\mathcal{D}$: Dataset, $\mathcal{G_D}$: Dictionary rules, ${\tau}_{p} , {\tau}_{\Delta} , {\tau}_{r} , {\tau}_{c}$: Thresholds
\Ensure $\mathcal{G_C}$: Combination rules

\Function{MineCombinationRules}{$\mathcal{D} , \mathcal{G_D}$}
	\Let {$\mathcal{{G_C}'}$} {$\{\}$}
	\For {$(X \rightarrow f)$ in $\mathcal{G_D}$} \Comment{$f : {w}_{0} {w}_{1} \cdots {w}_{ |f|-1 }$}
		\ForAll{$i,j ~~ s.t. ~~ 0 \le i < j \le |f|$}
			\If {$P(\mathcal{L}_{{w}_{i}^{j}} | {w}_{i}^{j}) > {\tau}_{p}$} \Comment{Polarity label $\mathcal{L}_{{w}_{i}^{j}} = \argmax_{\mathcal{X} \in \{\mathcal{N},\mathcal{P}\}}{P(\mathcal{X}| {w}_{i}^{j} )}$}
				\State {$r$: $X \rightarrow {w}_{0}^{ i } {L}_{{w}_{i}^{j}} {w}_{j}^{ |f| }$} \Comment{Non-terminal ${L}_{{w}_{i}^{j}} = \argmax_{{X} \in \{{N},{P}\}}{P(\mathcal{X}| {w}_{i}^{j} )} $}
				\If {$\mathcal{X} \neq \mathcal{L}_{{w}_{i}^{j}} $}
					\AddOne {$\#( r , negation)$}
				\ElsIf {$P(\mathcal{X} | f) > P(\mathcal{L}_{{w}_{i}^{j}} | {w}_{i}^{j}) + {\tau}_{\Delta} $}
					\AddOne {$\#( r , strengthen)$}
				\ElsIf {$P(\mathcal{X} | f) < P(\mathcal{L}_{{w}_{i}^{j}} | {w}_{i}^{j}) - {\tau}_{\Delta} $}
					\AddOne {$\#( r , weaken)$}
				\EndIf
				\State {add $r$ to $\mathcal{{G_C}'}$}
			\EndIf
		\EndFor
		\ForAll{$i_{0},j_{0},i_{1},j_{1} ~~ s.t. ~~ 0 \le i_{0} < j_{0}< i_{1} < j_{1} \le |f|$}
			\If {$P(\mathcal{L}_{{w}_{i_0}^{j_0}} | {w}_{i_0}^{j_0}) > {\tau}_{p}$~\textbf{and}~$P(\mathcal{L}_{{w}_{i_1}^{j_1}} | {w}_{i_1}^{j_1}) > {\tau}_{p}$}
				\State {$r$: $X \rightarrow {w}_{0}^{i_0} {L}_{{w}_{i_0}^{j_0}} {w}_{j_0}^{i_1} {L}_{{w}_{i_1}^{j_1}} {w}_{j_1}^{ |f| }$} \Comment{Replace ${w}_{i_0}^{j_0}$, ${w}_{i_1}^{j_1}$ with the non-terminals}
				\If {$\mathcal{L}_{{w}_{i_0}^{j_0}} \neq \mathcal{L}_{{w}_{i_1}^{j_1}} $}
					\AddOne {$\#( r , contrast )$}
				\EndIf
				\State {add $r$ to $\mathcal{{G_C}'}$}
			\EndIf
		\EndFor
	\EndFor
	\Let {$\mathcal{G_C}$} {$\{\}$}
	\For {$r$ in $\mathcal{{G_C}'}$}
		\If {$\#(r , \cdot) > {\tau}_{r}$ \textbf{and} $\underset {T}{\max}{\frac {\#(r , T)} {\#(r)}} > {\tau}_{c}$}
			\State {add $r$ to $\mathcal{G_C}$}
		\EndIf
	\EndFor
	\State \Return{$\mathcal{G_C}$}
\EndFunction
\end{algorithmic}
\end{algorithm}

\subsubsection{Polarity Model Learning} \label{sec:learn_computation_model}
As shown in Section~\ref{sec:polarity_model}, we define the polarity model to calculate the polarity probabilities using the sentiment grammar.
In this section, we present how to learn the parameters of the polarity model for the combination rules.

\begin{figure}[t]
\begin{center}
\includegraphics[width=0.97\textwidth]{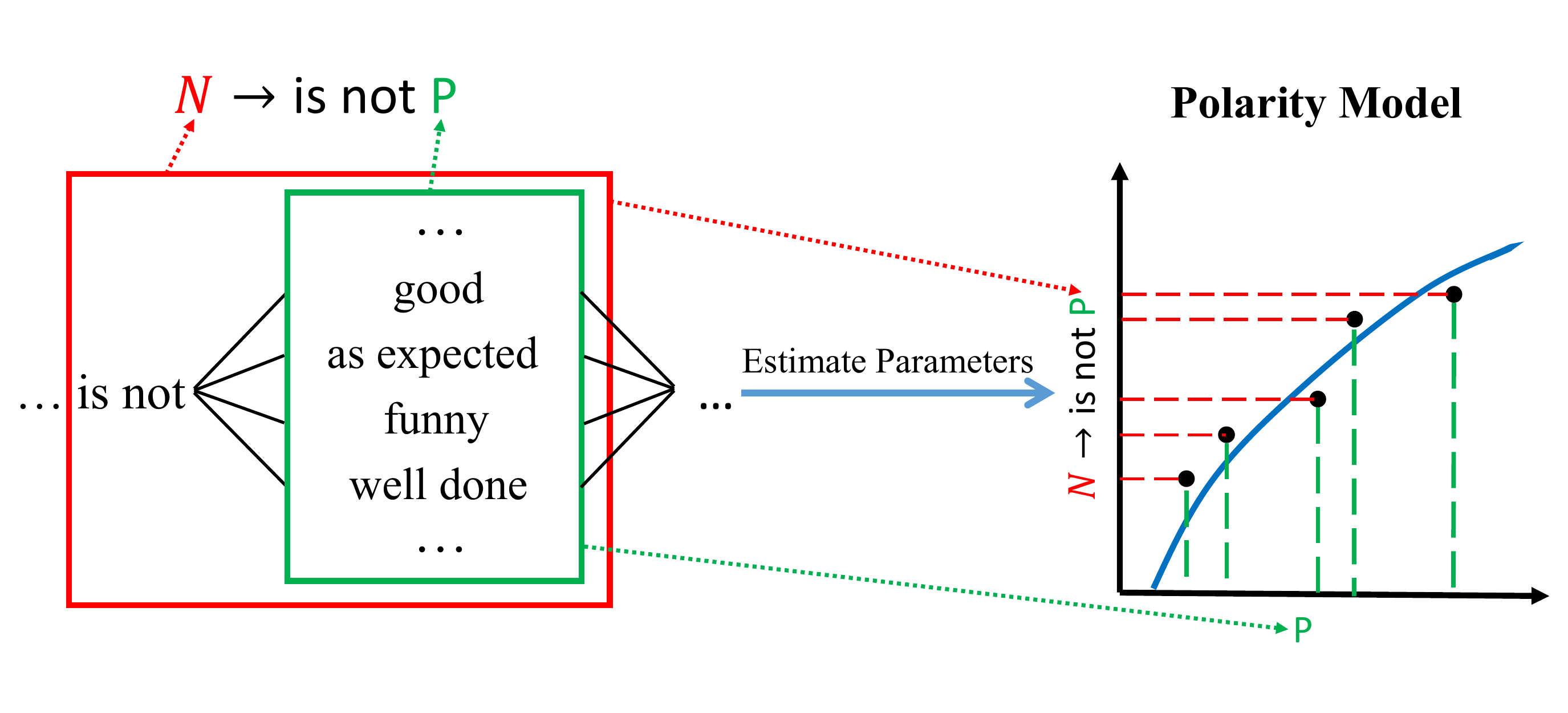}
\caption{We replace the subsequences with their polarity labels for frequent sentiment fragments. As shown in the above figure, we replace \textit{good}, \textit{as expected}, \textit{funny}, \textit{well done} with their polarity label $P$. Then we compare the polarity probabilities of sub-fragments with the whole fragments, such as \textit{good} and \textit{is not good}, to determine whether it is a negation rule, strengthen rule, or weaken rule. After obtaining the rule, we employ polarity probabilities of these compositional examples as training data to estimate parameters of the polarity model. In the above example, $\left( P(\mathcal{P}|\mbox{good}) , P(\mathcal{N}|\mbox{is~not~good}) \right)$, $\left( P(\mathcal{P}|\mbox{as expected}) , P(\mathcal{N}|\mbox{is~not~as~expected}) \right)$, $\left( P(\mathcal{P}|\mbox{funny}) , P(\mathcal{N}|\mbox{is~not~funny}) \right)$, $\left( P(\mathcal{P}|\mbox{well~done}) , P(\mathcal{N}|\mbox{is~not~well~done}) \right)$ are used to learn the polarity model for $N \rightarrow \mbox{is~not}~P$.}
\label{fig:computation_model}
\end{center}
\end{figure}

As shown in Figure~\ref{fig:computation_model}, we learn combination rules by replacing the subsequences of frequent sentiment fragments with their polarity labels. Both the replaced fragment and the whole fragment can be found in the dictionary rules, so their polarity probabilities have been estimated from data. We can employ them as our training examples to figure out how context changes the polarity of replaced fragment, and learn parameters of the polarity model.

We describe the polarity model in Section~\ref{sec:polarity_model}. To further simplify the notation, we denote the input vector $\mathbf{x} = (1 , P(\mathcal{X}_{1} | w_{i_1}^{j_1}) , \dots , P(\mathcal{X}_K | w_{i_K}^{j_K}))^{\mathsf{T}}$, and the response value as $y$.
Then we can rewrite Equation~\eqref{eq:logistic_model_func} as,
\begin{equation}
h_{\mathbf{\theta}}{(\mathbf{x})} = \frac{1}{1 + \exp\{- \mathbf{\theta}^{\mathsf{T}} \mathbf{x} \}}
\end{equation}
where $h_{\mathbf{\theta}}{(\mathbf{x})}$ is the polarity probability calculated by the polarity model, and $\mathbf{\theta} = (\mathbf{\theta}_0 , \mathbf{\theta}_1 , \dots , \mathbf{\theta}_K )^{\mathsf{T}}$ is the parameter vector.
Our goal is to estimate the parameter vector $\mathbf{\theta}$ of the polarity model.

We fit the model to minimize the sum of squared residuals between the predicted polarity probabilities and the values computed from data. We define the cost function as,
\begin{equation}
\begin{aligned}
\mathcal{J}(\mathbf{\theta})=\frac{1}{2}\sum_{m} {{(h_{\mathbf{\theta}}(\mathbf{x}^m) - y^m )}^{2}}
\end{aligned}
\end{equation}
where $\left( \mathbf{x}^m, y^m \right)$ is the $m$-th training instance.

The gradient descent algorithm is used to minimize the cost function $\mathcal{J}(\mathbf{\theta})$. The partial derivative of $\mathcal{J}(\mathbf{\theta})$ with respect to $\mathbf{\theta}_j$ is,
\begin{equation}
\begin{aligned}
\frac { \partial {\mathcal{J}(\mathbf{\theta})} }{ \partial {\mathbf{\theta}_j} } &=  \sum_{m}{\left( h_{\mathbf{\theta}}( \mathbf{x}^m ) - y^m \right) \frac { \partial { h_{\mathbf{\theta}}( \mathbf{x}^m ) } }{ \partial {\mathbf{\theta}_j} } } \\
&= \sum_{m}{\left( h_{\mathbf{\theta}}( \mathbf{x}^m ) - y^m \right)h_{\mathbf{\theta}} ( \mathbf{x}^m ) \left( 1-h_{\mathbf{\theta}}( \mathbf{x}^i ) \right) \frac { \partial { \mathbf{\theta}^{\mathsf{T}} \mathbf{x^m} } }{ \partial {\mathbf{\theta}_j} } } \\
&= \sum_{m}{\left( h_{\mathbf{\theta}}( \mathbf{x}^m ) - y^m \right)h_{\mathbf{\theta}} ( \mathbf{x}^m ) \left( 1-h_{\mathbf{\theta}}( \mathbf{x}^m ) \right) \mathbf{x}_j^m}
\end{aligned}
\end{equation}

We set the initial $\mathbf{\theta}$ as zeros, and start with it. We employ the Stochastic Gradient Descend algorithm to minimize the cost function. For the instance $(\mathbf{x},y)$, the parameters are updated using:
\begin{equation}
\label{eq:computation_model_update}
\begin{aligned}
{\mathbf{\theta}_j}^{(t+1)} &= {\mathbf{\theta}_j}^{(t)} - \alpha \left( \frac { \partial {\mathcal{J}(\mathbf{\theta})} }{ \partial {\mathbf{\theta}_j} } |_{\mathbf{\theta} = {\mathbf{\theta}}^{(t)} } \right) \\
&= {\mathbf{\theta}_j}^{(t)} - \alpha (h_{{\mathbf{\theta}}^{(t)}} ( \mathbf{x} )-y)h_{{\mathbf{\theta}}^{(t)}}( \mathbf{x} ) \left( 1-h_{{\mathbf{\theta}}^{(t)}} ( \mathbf{x} ) \right)\mathbf{x}_j
\end{aligned}
\end{equation}
where $\alpha$ is the learning rate, and it is set to $0.01$ in our experiments. We summarize the learning method in Algorithm~\ref{alg:computation_model_learning}. For each combination rule, we iteratively scan through the training examples $\left( \mathbf{x},y \right)$ in a random order, and update the parameters $\mathbf{\theta}$ according to Equation~\eqref{eq:computation_model_update}. The stopping condition is ${\left\| \mathbf{\theta}^{(t+1)}-\mathbf{\theta}^{(t)} \right\| }_{2}^{2} < \varepsilon$, which indicates the parameters become stable.

\begin{algorithm}[tb]
\caption{Polarity Model Learning Algorithm
\label{alg:computation_model_learning}}
\begin{algorithmic}[1]
\Require $\mathcal{G_C}$: Combination rules, $\varepsilon$: Stopping condition, $\alpha$: Learning rate
\Ensure $\mathbf{\theta}$: Parameters of the polarity model

\Function{EstimatePolarityModel}{$\mathcal{G_C}$}
	\ForAll {combination rule $r \in \mathcal{G_C}$}
		\Let {$\mathbf{\theta}^{(0)}$} {${(0,0,...,0)}^{\mathsf{T}}$}
		\Repeat
			\Let {$\left( \mathbf{x},y \right)$} {randomly select a training instance}
			\Let {${\mathbf{\theta}_j}^{(t+1)}$} {${\mathbf{\theta}_j}^{(t)} - \alpha (h_{{\mathbf{\theta}}^{(t)}} ( \mathbf{x} )-y)h_{{\mathbf{\theta}}^{(t)}}( \mathbf{x} ) \left( 1-h_{{\mathbf{\theta}}^{(t)}} ( \mathbf{x} ) \right)\mathbf{x}_j$}
			\Let{$t$}{$t+1$}
		\Until {${\left\| \mathbf{\theta}^{(t+1)}-\mathbf{\theta}^{(t)} \right\| }_{2}^{2} < \varepsilon$}
		\State {assign $\mathbf{\theta}^{(T)}$ as the parameters of the polarity model for rule $r$}
	\EndFor
\EndFunction
\end{algorithmic}
\end{algorithm}

\subsubsection{Summary of Grammar Learning Algorithm}
We summarize the grammar learning process in Algorithm~\ref{alg:fragment_rule_learning}, which learns the sentiment grammar in an iterative manner.

\begin{algorithm}[t]
\caption{Sentiment Grammar Learning
\label{alg:fragment_rule_learning}}
\begin{algorithmic}[1]
\Require $\mathcal{D}$: Dataset $\{ (s,\mathcal{L}_{s}) \}$, $T$: Maximum number of iteration \Comment{$\mathcal{L}_{s}$: Polarity~label~of~$s$}
\Ensure $\mathcal{G_D}$: Dictionary rules, $\mathcal{G_C}$: Combination rules
\Let {$\mathcal{G_C}$} {$\{\}$}
\Repeat
\Let{$\mathcal{G_D}$}{\Call{MineDictionaryRules}{$\mathcal{D} , \mathcal{G_C}$}} \Comment{Algorithm~\ref{alg:dictionary_rule_learning}}
\Let{$\mathcal{G_C}$}{\Call{MineCombinationRules}{$\mathcal{D} , \mathcal{G_D}$}} \Comment{Algorithm~\ref{alg:combination_rule_learning}}
\Until {iteration number exceeds $T$}
\State \Call{EstimatePolarityModel}{$\mathcal{G_C}$} \Comment{Algorithm~\ref{alg:computation_model_learning}}
\State \Return{$\mathcal{G_D} , \mathcal{G_C}$}
\end{algorithmic}
\end{algorithm}

We first learn the dictionary rules and their polarity probabilities by counting the frequencies in negative and positive classes. Only the fragments whose occurrence numbers are larger than the threshold $\tau_f$ are kept.
As mentioned in Section~\ref{sec:dictionary_rule_learning}, the context can essentially change the distribution of sentiment fragments. We take the combination rules into consideration to acquire more reasonable $\mathcal{G_D}$. In the first iteration, the set of combination rules is empty. Therefore, we have no information about compositionality to improve dictionary rule learning. The initial $\mathcal{G_D}$ contains some inaccurate sentiment distributions.
Next, we replace the subsequences of dictionary rules to their polarity labels, and generalize these sentiment fragments to the combination rules $\mathcal{G_C}$ as illustrated in Section~\ref{sec:combination_rule_learning}. At the same time, we can obtain their compositional types and learn parameters of the polarity model.
We iterate over the above two steps to obtain refined $\mathcal{G_D}$ and $\mathcal{G_C}$.

\section{Experimental Studies} \label{sec:experiment}
In this section, we describe experimental results on existing benchmark datasets with extensive comparisons with state-of-the-art sentiment classification methods. We also present the effects of different experimental settings in the proposed statistical sentiment parsing framework.

\subsection{Experiment Setup} \label{exp:setup}
We describe the datasets in Section~\ref{exp:datasets}, the experimental settings in Section~\ref{exp:settings}, and the methods used for comparison in Section~\ref{exp:campared_methods}.

\subsubsection{Datasets} \label{exp:datasets}
We conduct experiments on sentiment classification for sentence-level and phrase-level data. The sentence-level datasets contain user reviews and critic reviews from Rotten Tomatoes\footnote{http://www.rottentomatoes.com} and IMDB\footnote{http://www.imdb.com}. We balance the positive and negative instances in the training dataset to mitigate the problem of data imbalance.
Moreover, the Stanford Sentiment Treebank\footnote{http://nlp.stanford.edu/sentiment/treebank.html} contains polarity labels of all syntactically plausible phrases.
In addition, we use the MPQA\footnote{http://mpqa.cs.pitt.edu/corpora/mpqa\_corpus} dataset for the phrase-level task.
We describe these datasets as follows.

\textbf{RT-C}: 436,000 critic reviews from Rotten Tomatoes. It consists of 218,000 negative and 218,000 positive critic reviews. The average review length is 23.2 words. Critic reviews from Rotten Tomatoes contain a label (Rotten: Negative, Fresh: Positive) to indicate the polarity, which we use directly as the polarity label of corresponding review.

\textbf{PL05-C}: The sentence polarity dataset v1.0~\cite{Pang:2005} contains 5,331 positive and 5,331 negative snippets written by critics from Rotten Tomatoes. This dataset is widely used as the benchmark dataset in the sentence-level polarity classification task. The data source is the same as RT-C.

\textbf{SST}: The Stanford Sentiment Treebank~\cite{SocherEtAl2013:RNTN} is built upon PL05-C. The sentences are parsed to parse trees. Then, 215,154 syntactically plausible phrases are extracted and annotated by workers from Amazon Mechanical Turk. The experimental settings of positive/negative classification for sentences are the same as~\cite{SocherEtAl2013:RNTN}.

\textbf{RT-U}: 737,806 user reviews from Rotten Tomatoes. As we focus on sentence-level sentiment classification, we filter out user reviews that are longer than 200 characters. 
The average length of these short user reviews from Rotten Tomatoes is 15.4 words.
Following previous work on polarity classification, we use the review score to select highly polarized reviews. For the user reviews from Rotten Tomatoes, a negative review has a score $<$ 2.5 out of 5, and a positive review has a score $>$ 3.5 out of 5.

\textbf{IMDB-U}: 600,000 user reviews from IMDB. The user reviews in IMDB contain comments and short summaries (usually a sentence) to summarize the overall sentiment expressed in the reviews. We use the review summaries as the sentence-level reviews. The average length is 6.6 words.
For user reviews of IMDB, a negative review has a score $<$ 4 out of 10, and a positive review has a score $>$ 7 out of 10.

\textbf{C-TEST}: 2,000 labeled critic reviews sampled from RT-C. We use C-TEST as the testing dataset for RT-C. It should be mentioned that we exclude them from the training dataset (namely RT-C).

\textbf{U-TEST}: 2,000 manually labeled user reviews sampled from RT-U. 
User reviews often contain some noisy ratings compared to critic reviews. To eliminate the effect of noise, we sample 2,000 user reviews from RT-U, and annotate their polarity labels manually.
We use U-TEST as a testing dataset for RT-U and IMDB-U which are both user reviews.
It should be mentioned that we exclude them from the training dataset (namely RT-U).

\textbf{MPQA}: The opinion polarity subtask of the MPQA dataset~\cite{mpqa}. The authors manually annotate sentiment polarity labels for the expressions (i.e. sub-sentences) within a sentence.
We regard the expressions as short sentences in our experiments. There are 7,308 negative examples and 3,316 positive examples in this dataset. The average number of words per example is 3.1.

Table~\ref{table:dataset_size} shows the summary of these datasets, and all of them are publicly available at http://goo.gl/WxTdPf.

\begin{table}[tb]
\caption{Statistical information of datasets. \#Negative and \#Positive are the number of negative instances and positive instances, respectively. $l_{avg}$ is average length of sentences in the dataset, and $\left| V \right|$ is the vocabulary size.}
\begin{tabular*}{\textwidth}{l l l l l l}
\hline
Dataset & Size     & \#Negative & \#Positive	& $l_{avg}$	& $\left| V \right|$	\\ \hline
RT-C    & 436,000  & 218,000 	& 218,000 		& 23.2	& 136,006	\\
PL05-C	& 10,662   & 5,331   	& 5,331 		& 21.0	& 20,263	\\
SST		& 98,796   & 42,608   	& 56,188 		& 7.5	& 16,372	\\
RT-U 	& 737,806  & 368,903 	& 368,903 		& 15.4	& 138,815	\\
IMDB-U 	& 600,000  & 300,000 	& 300,000 		& 6.6	& 83,615	\\
MPQA	& 10,624   & 7,308   	& 3,316 		& 3.1	& 5,992		\\
\hline
\end{tabular*}
\label{table:dataset_size}
\end{table}

\subsubsection{Settings} \label{exp:settings}
To compare with other published results for PL05-C and MPQA, the training and testing regime (10-fold cross-validation) is the same as in~\cite{Pang:2005,tree-crf,rae2011}.
For SST, the regime is the same as in~\cite{SocherEtAl2013:RNTN}.
We use C-TEST as testing data for RT-C, and U-TEST as testing data for RT-U and IMDB-U.
There are a number of settings that have trade-offs in performance, computation, and the generalization power of our model. The best settings are chosen by a portion of training split data which serves as the validation set.
We provide the performance comparisons using different experimental settings in Section~\ref{exp:effect-of-settings}.

\textbf{Number of training examples}:
The size of training data has been widely recognized as one of the most important factors in machine learning based methods. Generally, using more data leads to better performance. By default, all the training data is used in our experiments. We use the same size of training data in different methods for fair comparisons.

\textbf{Number of training iterations ($T$)}:
We use AdaGrad~\cite{Duchi:2011:ASM:1953048.2021068} as the optimization algorithm in the learning process. 
The algorithm starts with randomly initialized parameters, and alternates between searching candidate sentiment trees and updating parameters of the ranking model. We treat one-pass scan of training data as an iteration.

\textbf{Beam size ($K$)}:
The beam size is used to make a trade-off between the search space and the computation cost. Moreover, an appropriate beam size can prune unfavorable candidates. We set $K=30$ in our experiments.

\textbf{Regularization ($\lambda$)}:
The regularization parameter $\lambda$ in Equation~\eqref{eq:obj_func} is used to avoid over-fitting. The value used in the experiments is $0.01$.

\textbf{Minimum fragment frequency}:
It is difficult to estimate reliable polarity probabilities when the fragment appears very few times. Hence, a minimum fragment frequency that is too small will introduce noise in the fragment learning process.
On the other hand, a large threshold will lose much useful information.
The minimum fragment frequency is chosen according to the size of the training dataset and the validation performance.
To be specific, we set this parameter as $4$ for RT-C, SST, RT-U, IMDB-U, and $2$ for PL05-C, MPQA.

\textbf{Maximum fragment length}:
High order n-grams are more precise and deterministic expressions than unigrams and bigrams. So it would be useful to employ long fragments to capture polarity information.
According to the experimental results, as the maximum fragment length increases, the accuracy of sentiment classification increases. The maximum fragment length is set to $7$ words in our experiments.

\subsubsection{Sentiment Classification Methods for Comparison} \label{exp:campared_methods}
We evaluate the proposed statistical sentiment parsing framework on the different datasets, and compare the results with some baselines and state-of-the-art sentiment classification methods described as follows.

\textbf{SVM-m}: Support Vector Machine (SVM) achieves good performance in the sentiment classification task~\cite{Pang:2005}. Though unigrams and bigrams are reported as the most effective features in existing work~\cite{Pang:2005}, we employ high-order n-gram ($1 \le n \le m$) features to conduct fair comparisons. Hereafter, $m$ has the same meaning. We employ LIBLINEAR~\cite{liblinear} in our experiments because it can well handle the high feature dimension and a large number of training examples. We try different hyper-parameters $C \in \{ {10}^{-2} , {10}^{-1} , 1 , 5 , 10 , 20 \}$ for SVM, and select $C$ on the validation set.

\textbf{MNB-m}: As indicated in~\cite{sidaw12acls}, Multinomial Na\"{i}ve Bayes (MNB) often outperforms SVM for sentence-level sentiment classification. We employ Laplace smoothing~\cite{Manning:2008:IIR:1394399} to tackle the zero probability problem. High order n-gram ($1 \le n \le m$) features are considered in the experiments.

\textbf{LM-m}: Language Model (LM) is a generative model calculating the probability of word sequences. It is used for sentiment analysis in~\cite{Cui:2006:CES:1597348.1597389}. Probability of generating sentence $s$ is calculated by $P(s)=\prod _{ i=0 }^{ \left| s \right| - 1  }{ P\left( { {w}_{i} }|{ {w}_{0}^{i-1} } \right)  }$, where ${w}_{0}^{ i-1 }$ denotes the word sequence ${w}_{0} \dots {w}_{i-1}$. We employ Good-Turing smoothing~\cite{GOOD01121953} to overcome sparsity when estimating the probability of high-order n-gram. We train language models on negative and positive sentences separately. For a sentence, its polarity is determined by comparing the probabilities calculated from the positive and negative language models. The unknown-word token is treated as a regular word (denoted by \textit{<UNK>}). SRI Language Modeling Toolkit~\cite{srilm} is used in our experiment.

\textbf{Voting-w/Rev}: This approach is proposed by~\namecite{choi2009adapting}, and is employed as a baseline in~\cite{tree-crf}. The polarity of a subjective sentence is decided by the voting of each phrase's prior polarity. The polarity of phrases that have odd numbers of negation phrases in their ancestors is reversed.
The results are reported by \namecite{tree-crf}.

\textbf{HardRule}: This baseline method is compared by~\namecite{tree-crf}. The polarity of a subjective sentence is deterministically decided based on rules, by considering the sentiment polarity of dependency subtrees. The polarity of a modifier is reversed if its head phrase has a negation word. The decision rules are applied from the leaf nodes to the root node in a dependency tree.
We use the results which are reported by \namecite{tree-crf}.

\textbf{Tree-CRF}: \namecite{tree-crf} present a dependency tree-based method employing conditional random fields with hidden variables. In this model, the polarity of each dependency subtree is represented by a hidden variable. The value of the hidden variable of the root node is identified as the polarity of the whole sentence.
The experimental results are reported by \namecite{tree-crf}.

\textbf{RAE-pretrain}: \namecite{rae2011} introduce a framework based on recursive autoencoders to learn vector space representations for multi-word phrases and predict sentiment distributions for sentences. We use the results with pre-trained word vectors learned on Wikipedia, which leads to better results compared to randomized word vectors. We directly compare the results with those in~\cite{rae2011}.

\textbf{MV-RNN}: \namecite{SocherEtAl2012:MVRNN} try to capture the compositional meaning of long phrases through matrix-vector recursive neural networks. This model assigns a vector and a matrix to every node in the parse tree. Matrices are regarded as operators, and vectors capture the meaning of phrases. The results are reported by~\namecite{SocherEtAl2012:MVRNN} and~\namecite{SocherEtAl2013:RNTN}.

\textbf{s.parser-LongMatch}: The longest matching rules are employed in the decoding process. In other words, the derivations that contain the fewest rules are used for all text spans. In addition, the dictionary rules are preferred to the combination rules if both of them match the same text span. The dynamic programming algorithm is used in the implementation.

\textbf{s.parser-w/oComb}: Our method without using the combination rules (such as ${N} \rightarrow \mbox{not}~P$) learned from data.

\subsection{Results of Sentiment Classification} \label{exp:result_overall}

\def\tableempiricalresultswidth{1.3cm}
\begin{table}[tb]
\caption{Sentiment classification results on different datasets. The top three methods are in \textbf{bold} and the best is also \underline{\textbf{underlined}}. SVM-m: Support Vector Machine. MNB-m: Multinomial Na\"{i}ve Bayes. LM-m: Language Model. Voting-w/Rev: Voting with negation rules. HardRule: Rule based method on dependency tree. Tree-CRF: Dependency tree-based method employing conditional random fields. RAE-pretrain: Recursive autoencoders with pre-trained word vectors. MV-RNN: Matrix-vector recursive neural network. s.parser-LongMatch: The longest matching rules are used. s.parser-w/oComb: Without using the combination rules. s.parser: Our method. Some of results are missing (indicated by ``-'') in the table as there is no publicly available implementation or they are hard to scale up.}
\begin{tabular*}{\textwidth}{l L{\tableempiricalresultswidth} L{\tableempiricalresultswidth} L{\tableempiricalresultswidth} L{\tableempiricalresultswidth} L{\tableempiricalresultswidth} L{\tableempiricalresultswidth}}
\hline
Method 			   & RT-C & PL05-C 		& SST 			& RT-U 			& IMDB-U 		& MPQA \\ \hline
SVM-1		       & 80.3 & 76.3   		& 81.1 			& 88.5 			& 84.9   		& 85.1 \\
SVM-2     & \textbf{83.0} & 77.4   		& 81.3			& 88.9 			& 86.8 	 		& 85.3 \\
SVM-3     & \textbf{83.1} & 77.0   		& 81.2			& 89.7			& \textbf{87.2} & 85.5 \\
SVM-4		       & 81.5 & 76.9   		& 80.9			& \textbf{89.8}	& 87.0 	 		& 85.6 \\
SVM-5		       & 81.7 & 76.8   		& 80.8			& 89.3 			& 87.0 	 		& 85.6 \\ \hline
MNB-1		       & 79.6 & 78.0   		& 82.6			& 83.3 			& 82.7   		& 85.0 \\
MNB-2			   & 82.0 & \textbf{78.8} & \textbf{83.3}	& 87.5 		& 85.6 	 		& 85.0 \\
MNB-3  			   & 82.2 & 78.4   		& \textbf{82.9}		& 88.6 		& 84.6 	 		& 85.0 \\
MNB-4  			   & 81.8 & 78.2   		& 82.6			& 88.2 			& 83.1 			& 85.1 \\
MNB-5  		 	   & 81.7 & 78.1   		& 82.4			& 88.1 			& 82.5 			& 85.1 \\ \hline
LM-1   			   & 77.6 & 75.1   		& 80.9			& 87.6 			& 81.8 			& 64.0 \\
LM-2 			   & 78.0 & 74.1   		& 78.4 			& 89.0 			& 85.8 			& 71.4 \\
LM-3 			   & 77.3 & 74.2   		& 78.3			& 89.3 			& \textbf{87.1}	& 71.1 \\
LM-4 			   & 77.2 & 73.0  		& 78.3			& 89.6 			& 87.0 			& 71.1 \\
LM-5 			   & 77.0 & 72.9   		& 78.2			& \textbf{90.0} & \textbf{87.1}	& 71.1 \\ \hline
Voting-w/Rev 	   & -	  & 63.1   		& -   		 	& -   		 	& -    	 		& 81.7 \\
HardRule 		   & -	  & 62.9		& -   		 	& -   		 	& -    	 		& 81.8 \\ \hline
Tree-CRF		   & -	  & 77.3		& -   		 	& -   		 	& -    	 		& \textbf{86.1} \\
RAE-pretrain   & -	  & 77.7		& -				& -    			& - 	 		& \underline{\textbf{86.4}} \\
MV-RNN 		       & -	& \textbf{79.0} & \textbf{82.9} & -    			& - 	 		& - \\ \hline
s.parser-LongMatch & 82.8	& 78.6		& 82.5			& 89.4   		& 86.9 	 		& 85.7 \\
s.parser-w/oComb   & 82.6 & 78.3		& 82.4			& 89.0 			& 86.4	 		& 85.5 \\
s.parser		   & \underline{\textbf{85.1}} & \underline{\textbf{79.5}} & \underline{\textbf{84.7}} & \underline{\textbf{91.5}} & \underline{\textbf{89.3}} & \textbf{86.2} \\ \hline
\end{tabular*}
\label{table:empirical_results}
\end{table}

We present the experimental results of the sentiment classification methods on the different datasets in Table~\ref{table:empirical_results}. The top three methods on each dataset are in bold, and the best methods are also underlined.
The experimental results show that s.parser achieves better performances than other methods on most datasets. 

The datasets RT-C, PL05-C, and SST are critic reviews.
On RT-C, the accuracy of s.parser increases by 2\%, 2.9\%, and 7.1\% from the best results of SVM, MNB, and LM, respectively.
On PL05-C, the accuracy of s.parser also rises by 2.1\%, 0.7\%, and 4.4\% from the best results of SVM, MNB, and LM, respectively. Comparing to Voting-w/Rev and HardRule, s.parser outperforms them by 16.4\% and 16.6\%. The results indicate that our method significantly outperforms the baselines which use manual rules, as rule-based methods lack a probabilistic way to model the compositionality of context. Furthermore, s.parser achieves an accuracy improvement rate of 2.2\%, 1.8\%, and 0.5\% over Tree-CRF, RAE-pretrain, and MV-RNN, respectively.
On SST, s.parser outperforms SVM, MNB, and LM by 3.4\%, 1.4\%, and 3.8\%, respectively. The performance is better than MV-RNN with an improvement rate of 1.8\%. Moreover, the result is comparable to the 85.4\% obtained by recursive neural tensor networks~\cite{SocherEtAl2013:RNTN} without depending on syntactic parsing results.

On the user review datasets RT-U and IMDB-U, our method also achieves the best results.
More specifically, on the dataset RT-U, s.parser outperforms the best results of SVM, MNB, and LM by 1.7\%, 2.9\%, and 1.5\%, respectively.
On the dataset IMDB-U, our method brings an improved accuracy rate by 2.1\%, 3.7\%, and 2.2\% over SVM, MNB, and LM, respectively. 
We find that MNB performs better than SVM and LM on the critics review datasets RT-C and PL05-C. Also, SVM and LM achieve better results on the user review datasets RT-U and IMDB-U. The s.parser is more robust for the different genres of datasets.

On the dataset MPQA, the accuracy of s.parser increases by 0.5\%, 1.1\%, and 14.8\% from the best results of SVM, MNB, and LM, respectively. Compared to Voting-w/Rev and HardRule, s.parser achieves 4.5\% and 4.4\% improvements over them. As illustrated in Table~\ref{table:dataset_size}, the size and length of sentences in MPQA are much smaller than those in the other four datasets. The RAE-pretrain achieves better results than other methods on this dataset, because the word embeddings pre-trained on Wikipedia can leverage smoothing to relieve the sparsity problem in MPQA. If we do not use any external resources (i.e. Wikipedia), the accuracy of RAE on MPQA is 85.7\% which is lower than Tree-CRF and s.parser.
The results indicate that s.parser achieves the best result if no external resource is employed.

In addition, we compare to the results of s.parser-LongMatch and s.parser-w/oComb. The s.parser-LongMatch utilizes the dictionary rules and combination rules in the longest matching manner, while s.parser-w/oComb removes the combination rules in the parsing process.
Compared with the results of s.parser, we find that both the ranking model and the combination rules play a positive role in the model. The ranking model learns to score parse trees by assigning larger weights to the rules that tend to obtain correct labels. Also, the combination rules generalize these dictionary rules to deal with the sentiment compositionality in a symbolic way, which enables the model to process unseen phrases.
Furthermore, s.parser-LongMatch achieves better results than s.parser-w/oComb. This indicates that the effects of the combination rules are more pronounced than the ranking model.

The bag-of-words classifiers work well for long documents relying on sentiment words that appear many times in a document. The redundancy characteristics provide strong evidence for sentiment classification. Even though some phrases of a document are not estimated accurately, it can still result in a correct polarity label. However, for short text, such as a sentence, the compositionality plays an important role in sentiment classification. 
Tree-CRF, MV-RNN and s.parser take compositionality into consideration in different ways, and they achieve significant improvements over SVM, MNB, and LM.
We also find that the high order n-grams contribute to classification accuracy on most of the datasets, but they harm the accuracy of LM on PL05-C. The high-order n-grams can partially solve compositionality in a brute-force way.

\subsection{Effect of Training Data Size} \label{exp:effect-of-data-size}
We further investigate the effect of the size of training data for different sentiment classification methods. This is meaningful as the number of the publicly available reviews is increasing dramatically nowadays. The methods that can take advantage of more training data will be even more useful in practice.

\begin{figure}[tb]
\begin{center}
\includegraphics[width=0.58\textwidth]{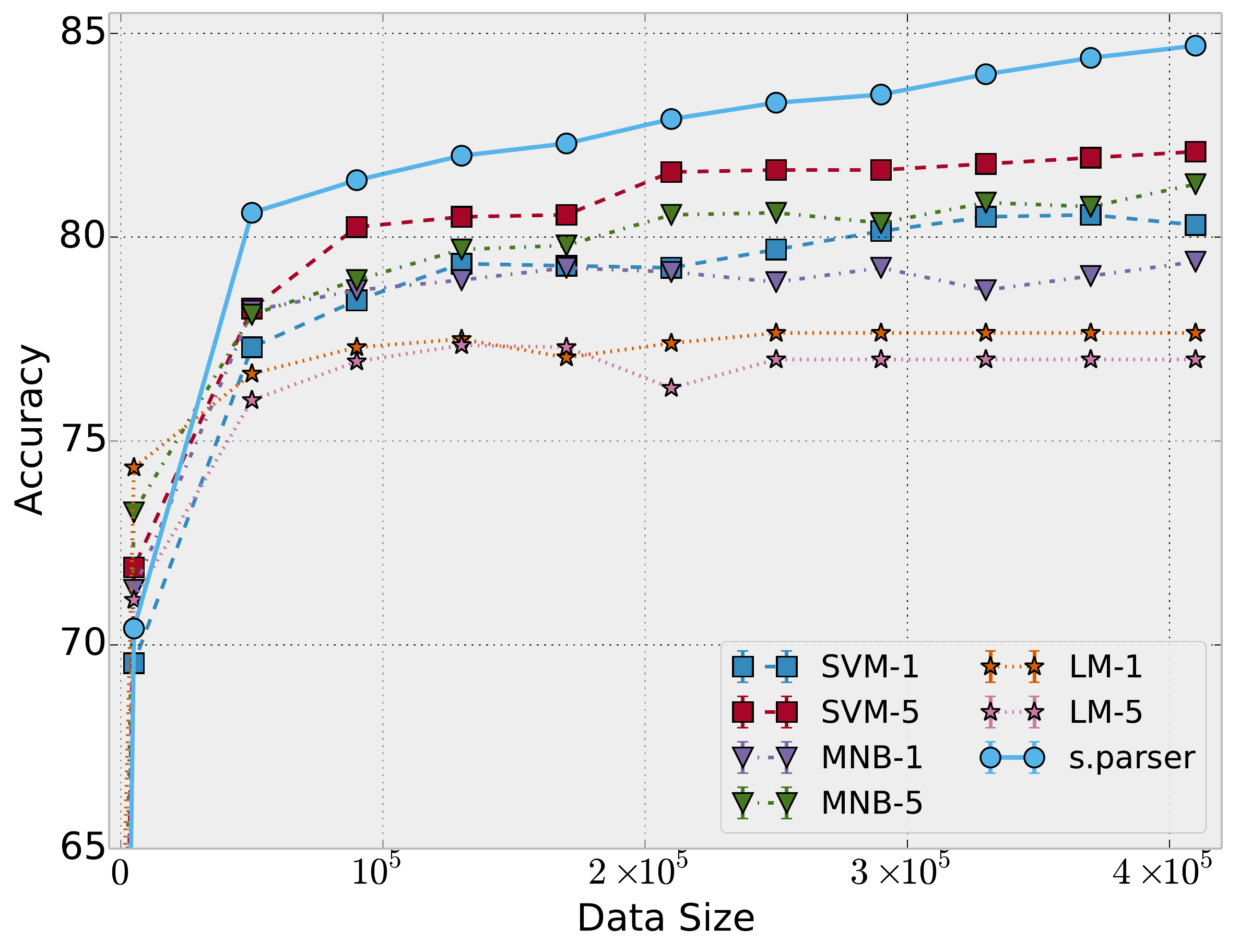}
\caption{The curves show the test accuracy as the number of training examples increases. Our method s.parser significantly outperforms the other methods, which indicates s.parser can leverage data more effectively and benefit more from larger data.}
\label{fig:setting:data_size}
\end{center}
\end{figure}

We report the results of s.parser compared with SVM, MNB, and LM on the dataset RT-C using different training data size. 
In order to make the figure clear, we only present the results of SVM/MNB/LM-1/5 here.
As shown in Figure~\ref{fig:setting:data_size}, we find that the size of training data plays an important role for all these sentiment classification methods. The basic conclusion is that the performances of all the methods rise as the data size increases.
To be specific, when the size is small, the performances ascend sharply. It meets our intuition that the size of data is the key factor when the size is relative small.
When the size of data is larger, the growth of accuracy becomes slower. The performances of the baseline methods start to converge after the data size is larger than 200,000.
The comparisons illustrate that s.parser significantly outperforms these baselines. And the performance of s.parser still becomes better when the data size increases. The convergence of s.parser's performance is slower than the others.
It indicates that s.parser leverages data more effectively and benefits more from a larger dataset. With more training data, s.parser learns more dictionary rules and combination rules. These rules enhance the generalization ability of our model. Furthermore, it estimates more reliable parameters for the polarity model and ranking model. In contrast, the bag-of-words based approaches (such as SVM, MNB, and LM) cannot make full use of high-order information in the dataset. The generalization ability of the combination rules of s.parser leads to better performance, and take advantage of larger data.
It should be noted that there are similar trends on other datasets.

%
%

\subsection{Effect of Experimental Settings} \label{exp:effect-of-settings}
In this section, we investigate the effects of different experimental settings. We show the results on the dataset RT-C by only changing a factor and fixing the others.

\begin{figure}[tb]
\begin{center}
\subfloat[Effect of minimum fragment frequency\label{fig:setting:fragment:a}]{%
	\includegraphics[width=0.47\textwidth]{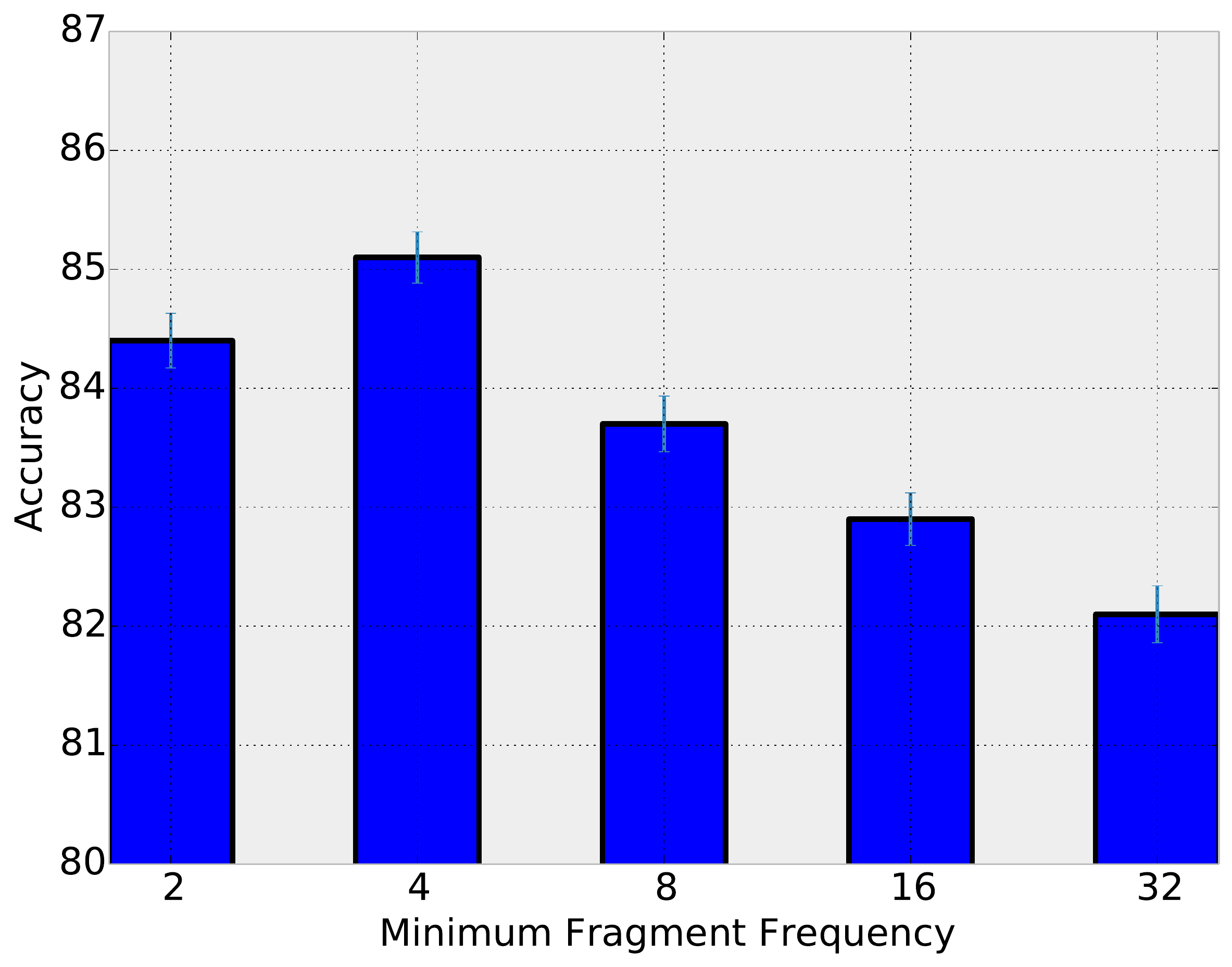}
}
\hfill
\subfloat[Effect of maximum fragment length\label{fig:setting:fragment:b}]{%
	\includegraphics[width=0.47\textwidth]{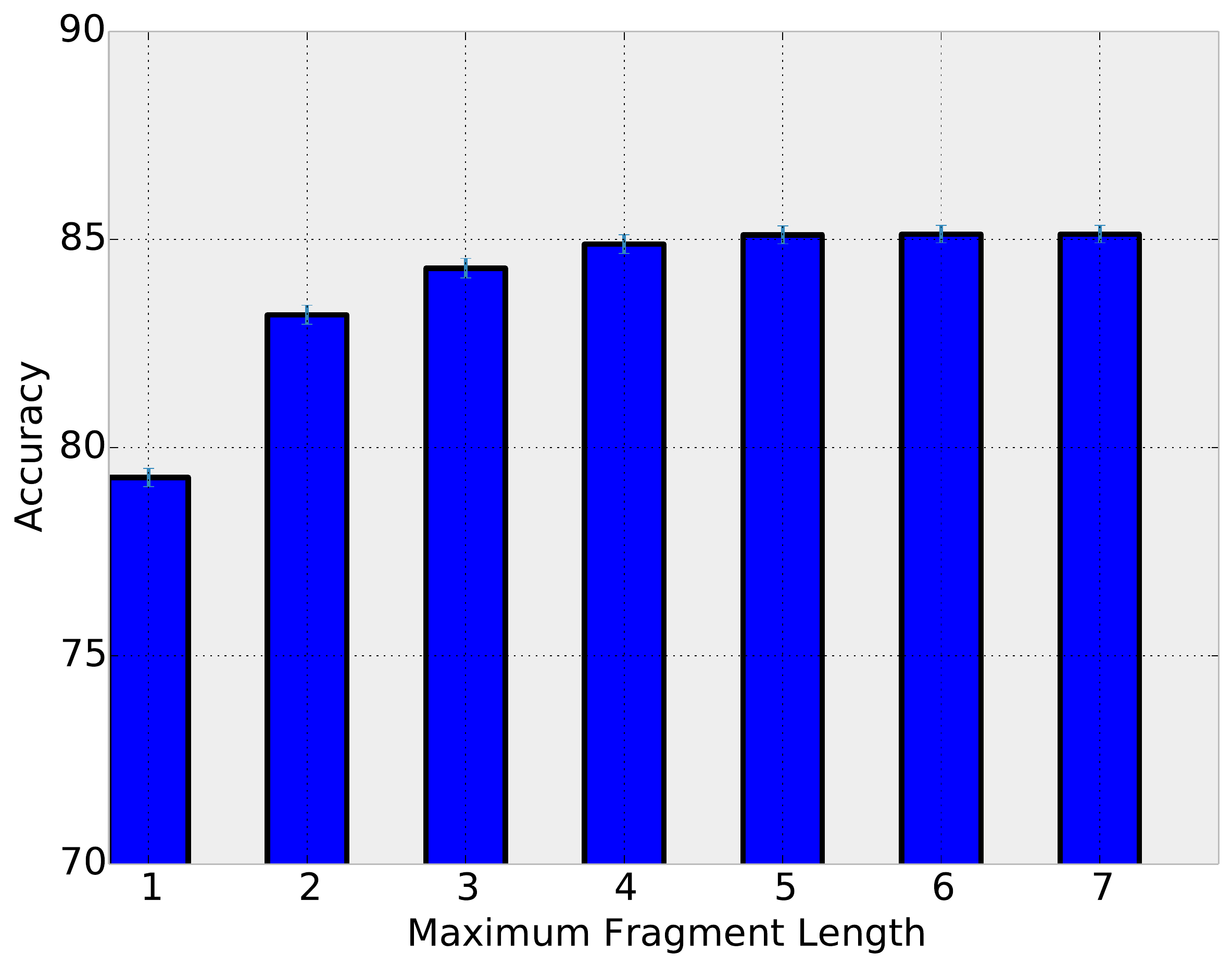}
}
\caption{(a) When the minimum fragment frequency is small, noise is introduced in the fragment learning process. On the other hand, too large threshold loses useful information. (b) As the maximum fragment length increases, the accuracy increases monotonically. It indicates that long fragments are useful for our method.}
\label{fig:setting:fragment}
\end{center}
\end{figure}

Figure~\ref{fig:setting:fragment} shows the effect of minimum fragment frequency, and maximum fragment length.
Specifically, Figure~\ref{fig:setting:fragment:a} indicates that a minimum fragment frequency that is too small will introduce noise, and it is difficult to estimate reliable polarity probabilities for infrequent fragments. However, a minimum fragment frequency that is too large will discard too much useful information.
As shown in Figure~\ref{fig:setting:fragment:b}, we find that accuracy increases as the maximum fragment length increases. The results illustrate that the large maximum fragment length is helpful for s.parser. We can learn more combination rules with a larger maximum fragment length, and long dictionary rules capture more precise expressions than unigrams. This conclusion is the same as that in Section~\ref{exp:result_overall}.

\begin{figure}[t]
\begin{center}
\subfloat[Effect of regularization\label{fig:parsing_setting:b}]{%
	\includegraphics[width=0.46\textwidth]{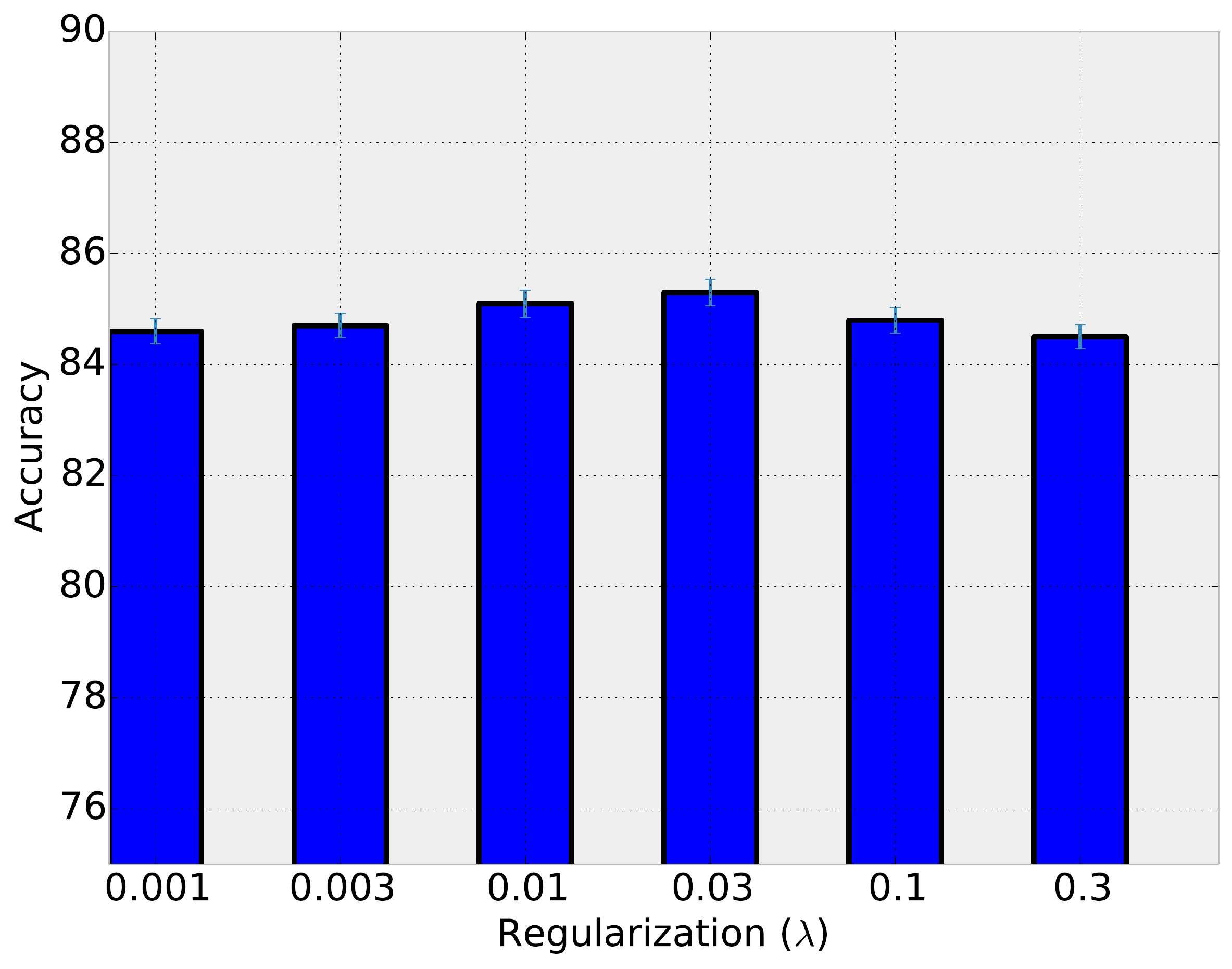}
}
\hfill
\subfloat[Effect of beam size\label{fig:parsing_setting:c}]{%
	\includegraphics[width=0.50\textwidth]{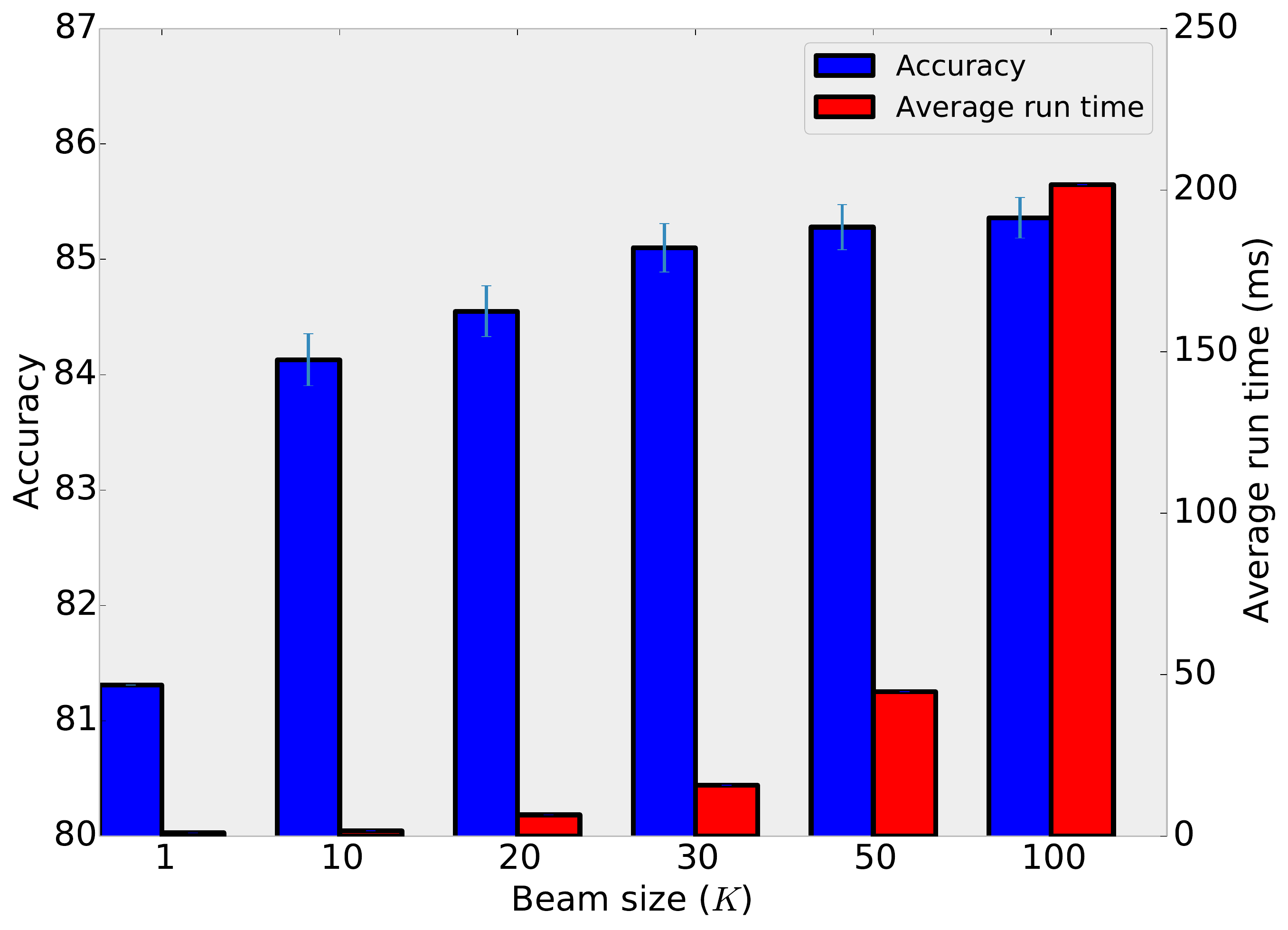}
}
\caption{
(a) The test accuracy is relatively insensitive to the regularization parameter $\lambda$ in Equation~\eqref{eq:obj_func}. (b) As the beam size $K$ increases, the test accuracy increases, however, the computation costs also become more expensive. When $K=1$, the optimization algorithm cannot learn any weights.}
\label{fig:parsing_setting}
\end{center}
\end{figure}

As shown in Figure~\ref{fig:parsing_setting}, we also investigate how the training iteration, regularization, and beam size affect the results.
As shown in Figure~\ref{fig:parsing_setting:b}, we try a wide range of regularization parameters $\lambda$ in Equation~\eqref{eq:obj_func}. The results indicate that it is insensitive to the choice of $\lambda$.
Figure~\ref{fig:parsing_setting:c} shows the effects of different beam size $K$ in the search process. When beam size $K=1$, the optimization algorithm cannot learn the weights. In this case, the decoding process is to select one search path randomly, and compute its polarity probabilities. The results become better as the beam size $K$ increases. On the other hand, the computation costs are more expensive. The proper beam size $K$ can prune some candidates to speed up the search procedure. It should be noted that the sentence length also effects the run time.

\subsection{Results of Grammar Learning} \label{exp:result_fragment}
The sentiment grammar plays a central role in the statistical sentiment parsing framework. It is obvious that the accuracy of s.parser relies on the quality of the automatically learned sentiment grammar. The quality can be implicitly evaluated by the accuracy of sentiment classification results as we have shown in previous sections. However, there is no straightforward way to explicitly evaluate the quality of the learned grammar. In this section, we will provide several case studies of the learned dictionary rules and combination rules to further illustrate the results of the sentiment grammar learning process as detailed in Section~\ref{sec:grammar_learning}.

To start with, we report the total number of dictionary rules and combination rules learned from the datasets.
As shown in Table~\ref{table:fragment_count}, the results indicate that we can learn more dictionary rules and combination rules from the larger datasets. Although we learn more dictionary rules from RT-C than from IMDB-U, the number of combination rules learned from RT-C is less than from IMDB-U. It indicates that the language usage of RT-C is more diverse than of IMDB-U. For SST, more rules are learned due to its constituent-level annotations.

\begin{table}[tb]
\caption{Number of rules learned from different datasets. $\tau_{f}$ represents minimum fragment frequency, $|\mathcal{G_D}|$ represents total number of dictionary rules, and $|\mathcal{G_C}|$ is the total number of combination rules.}
\begin{tabular*}{\textwidth}{l l l l}
\hline
Dataset	& $\tau_{f}$	& $|\mathcal{G_D}|$	& $|\mathcal{G_C}|$	\\ \hline
RT-C	& 4				& 758,723			& 952 \\
PL05-C	& 2				& 44,101			& 139 \\
SST		& 4				& 336,695			& 751 \\
RT-U	& 4				& 831,893			& 2,003 \\
IMDB-U	& 4				& 249,718			& 1,014 \\
MPQA	& 2				& 6,146				& 21 \\
\hline
\end{tabular*}
\label{table:fragment_count}
\end{table}

\begin{figure}[tb]
\begin{center}
\subfloat[Effect of minimum fragment frequency $\log_{2}(\tau_{f})$ \label{fig:fragment:a}]{%
	\includegraphics[width=0.45\textwidth]{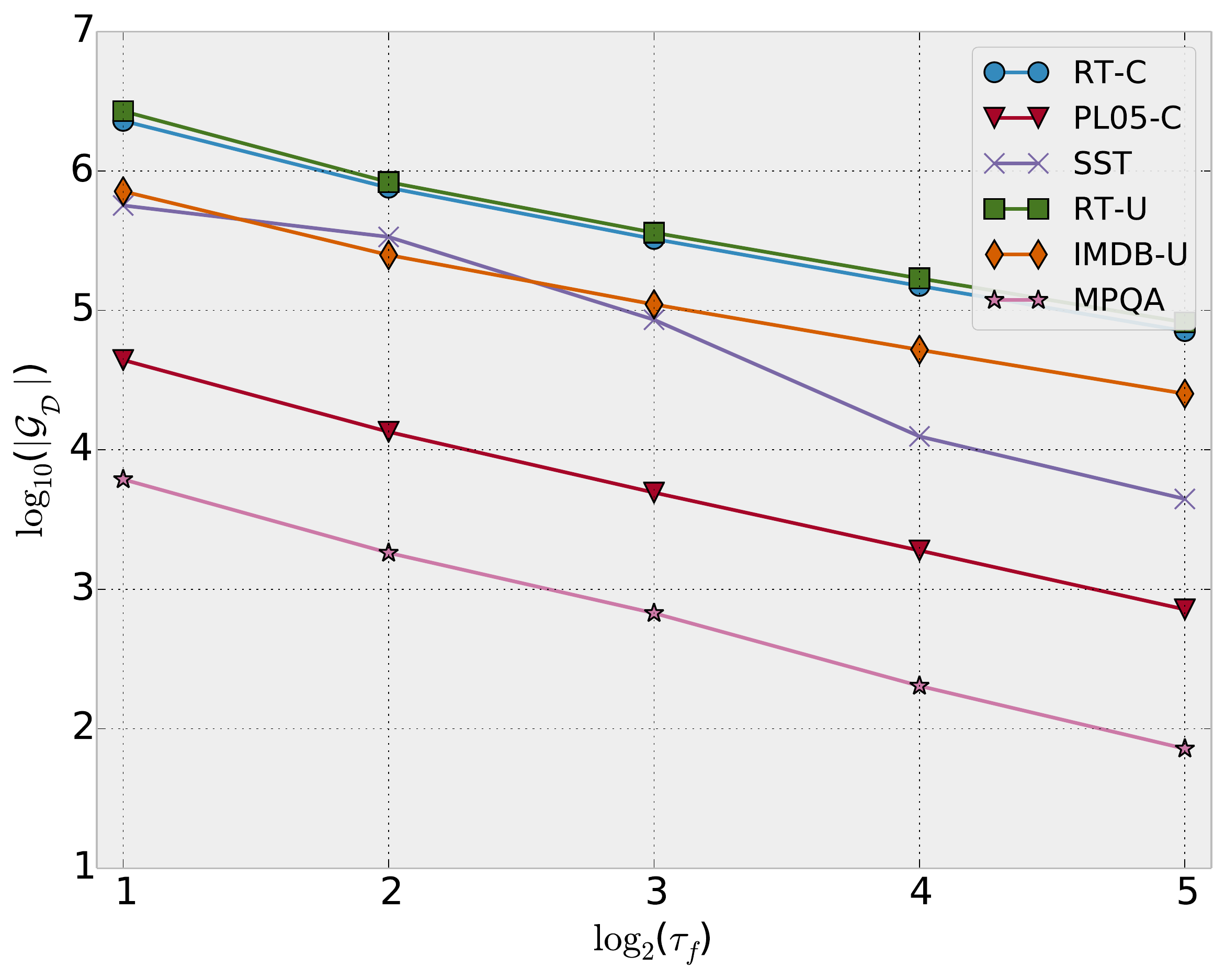}
}
\hfill
\subfloat[Cumulative distribution of dictionary rule length $l$\label{fig:fragment:b}]{%
	\includegraphics[width=0.45\textwidth]{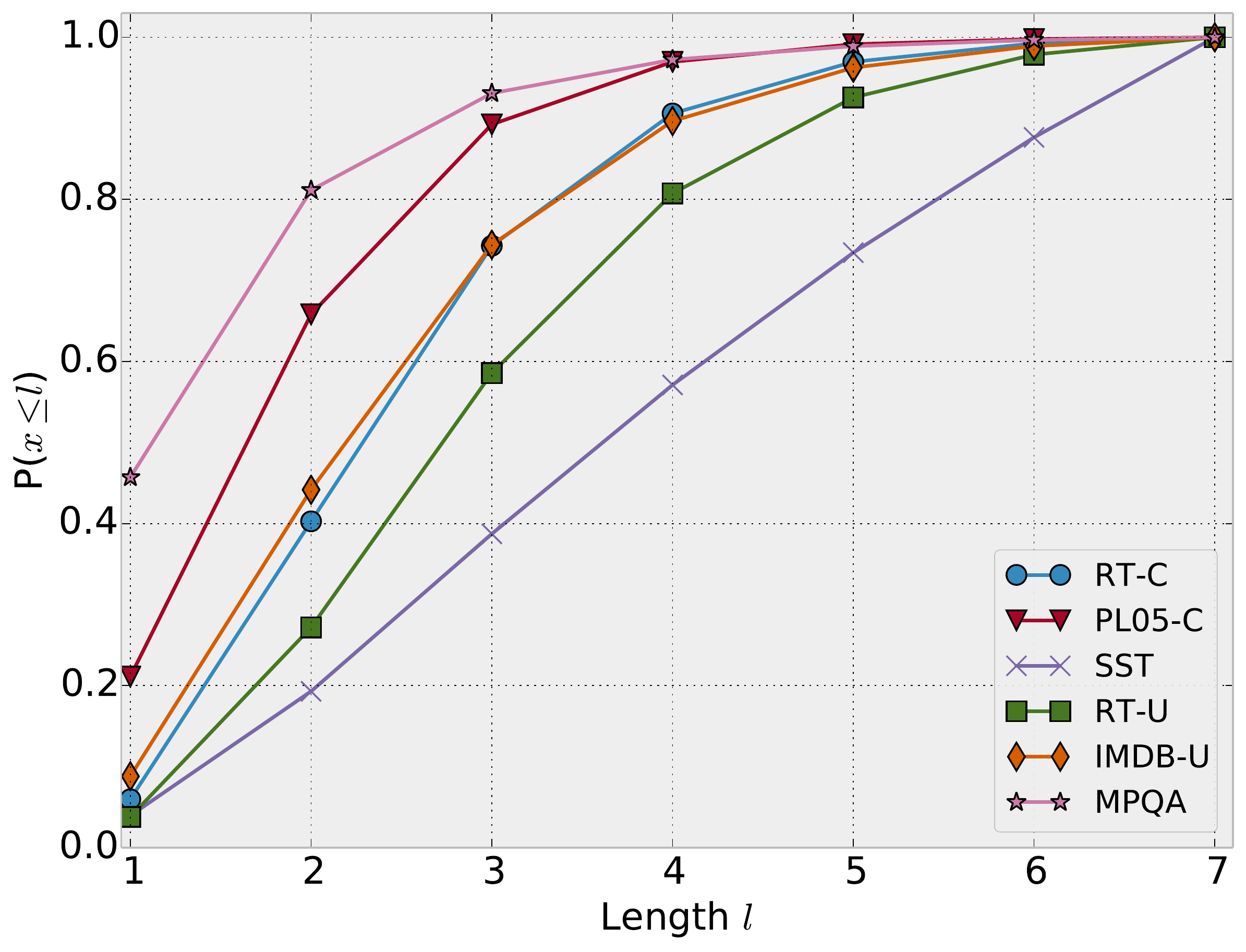}
}
\caption{(a) We choose $\tau_{f}=2,4,8,16,32$, and plot $\log_{10}(|\mathcal{G_D}|)$-$\log_{2}(\tau_{f})$ graph to show the effects of $\tau_{f}$ for total number of dictionary rules $|\mathcal{G_D}|$. The results (except SST) follow a power law distribution. (b) The cumulative distribution of dictionary rule length $l$ indicates that most dictionary rules are short ones.}
\label{fig:fragment}
\end{center}
\end{figure}

Furthermore, we explore how the minimum fragment frequency $\tau_{f}$ affects the number of dictionary rules, and present the distribution of dictionary rule length.
As illustrated in Figure~\ref{fig:fragment:a}, we find that the relation between total number of dictionary rules $|\mathcal{G_D}|$ and minimum fragment frequency $\tau_{f}$ obeys the power law, i.e., the $\log_{10}(|\mathcal{G_D}|) - \log_{2}(\tau_{f})$ graph takes a linear form. It indicates that most of the fragments appear few times, and only some of them appear frequently.
Notably, all the syntactically plausible phrases of SST are annotated, so its distribution is different from the other sentence-level datasets.
Figure~\ref{fig:fragment:b} shows the cumulative distribution of dictionary rule length $l$. It presents most dictionary rules are short ones. For all datasets except SST, more than 80\% of dictionary rules are shorter than five words. The length distributions of datasets RT-C and IMDB-U are similar, while we obtain more high order n-grams from RT-U and SST.

We further investigate the effect of context for dictionary rule learning. Table~\ref{table:fragment_comparison} shows some dictionary rules with polarity probabilities learned by our method and naive counting on RT-C. We notice that if we count the fragment occurrence number directly, some polarities of fragments are learned incorrectly. This is caused by the effect of context as described in Section~\ref{sec:dictionary_rule_learning}. By taking the context into consideration, we obtain more reasonable polarity probabilities of dictionary rules. Our dictionary rule learning method takes compositionality into consideration, i.e. we skip the count if there exist some negation indicators outside the phrase. This constraint tries to ensure that the polarity of fragment is the same as the whole sentence. As shown in the results, the polarity probabilities learned by our method are more reasonable and meet people's intuitions.
However, there are also some negative examples caused by ``false subjective''. For instance, the neutral phrase ``to pay it'' tends to appear in negative sentences, and it is learned as a negative phrase. This makes sense for the data distribution, while it may lead to the mismatch for the combination rules.

\begin{table}[t]
\caption{Comparing our dictionary rule learning method with naive counting. The dictionary rules which are assigned different polarities by these two methods are presented. $\mathcal{N}$ represents negative, and $\mathcal{P}$ represents positive. The polarity probabilities of fragments are shown in this table, and they demonstrate our method learns more intuitive results than counting directly.}
\begin{tabular*}{\textwidth}{l l l l l l l l}
\hline
\multirow{2}{*}{Fragment} & \multicolumn{3}{c}{Naive Count} & & \multicolumn{3}{c}{s.parser} \\
\cline{2-4}
\cline{6-8}
& $\mathcal{N}$ & $\mathcal{P}$ & Polarity & & $\mathcal{N}$ & $\mathcal{P}$ & Polarity \\ \hline
are fun             & 0.54 & 0.46 & $\mathcal{N}$ & & 0.11 & 0.89 & $\mathcal{P}$  \\
a very good movie 	& 0.61 & 0.39 & $\mathcal{N}$ & & 0.19 & 0.81 & $\mathcal{P}$  \\
looks gorgeous		& 0.56 & 0.44 & $\mathcal{N}$ & & 0.17 & 0.83 & $\mathcal{P}$  \\
to enjoy the movies	& 0.53 & 0.47 & $\mathcal{N}$ & & 0.14 & 0.86 & $\mathcal{P}$   \\
is corny			& 0.43 & 0.57 & $\mathcal{P}$ & & 0.83 & 0.17 & $\mathcal{N}$   \\
' s flawed			& 0.32 & 0.68 & $\mathcal{P}$ & & 0.63 & 0.37 & $\mathcal{N}$   \\
a difficult film to	& 0.43 & 0.57 & $\mathcal{P}$ & & 0.67 & 0.33 & $\mathcal{N}$   \\
disappoint			& 0.39 & 0.61 & $\mathcal{P}$ & & 0.77 & 0.23 & $\mathcal{N}$  \\
\hline
\end{tabular*}
\label{table:fragment_comparison}
\end{table}

\begin{figure}[t]
\begin{center}
\subfloat[$N \rightarrow \mbox{i~do~not}~P$\label{fig:experiment_computation_model:a}]{%
	\includegraphics[width=0.47\textwidth]{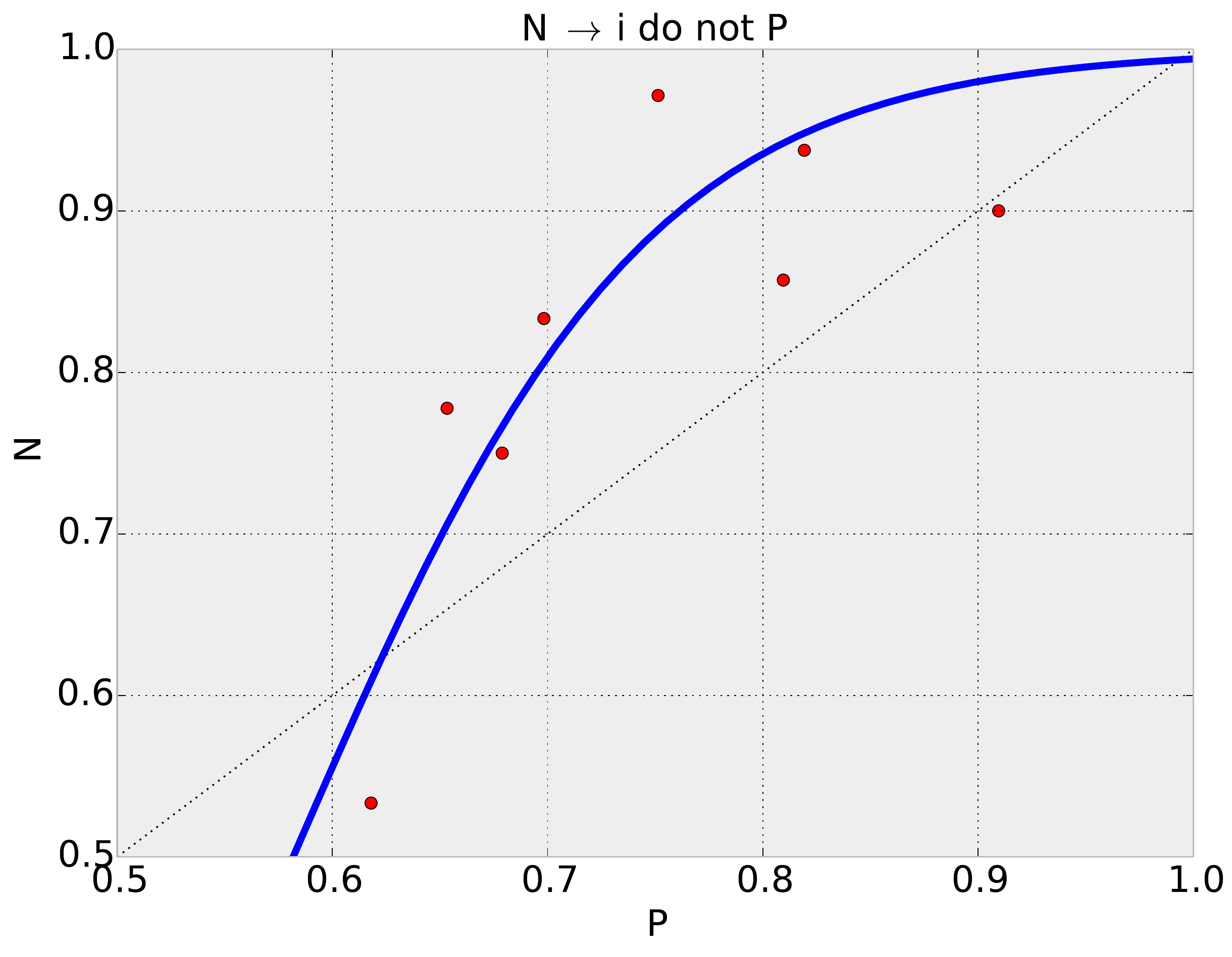}
}
\hfill
\subfloat[$N \rightarrow \mbox{is~not}~P .$\label{fig:experiment_computation_model:b}]{%
	\includegraphics[width=0.47\textwidth]{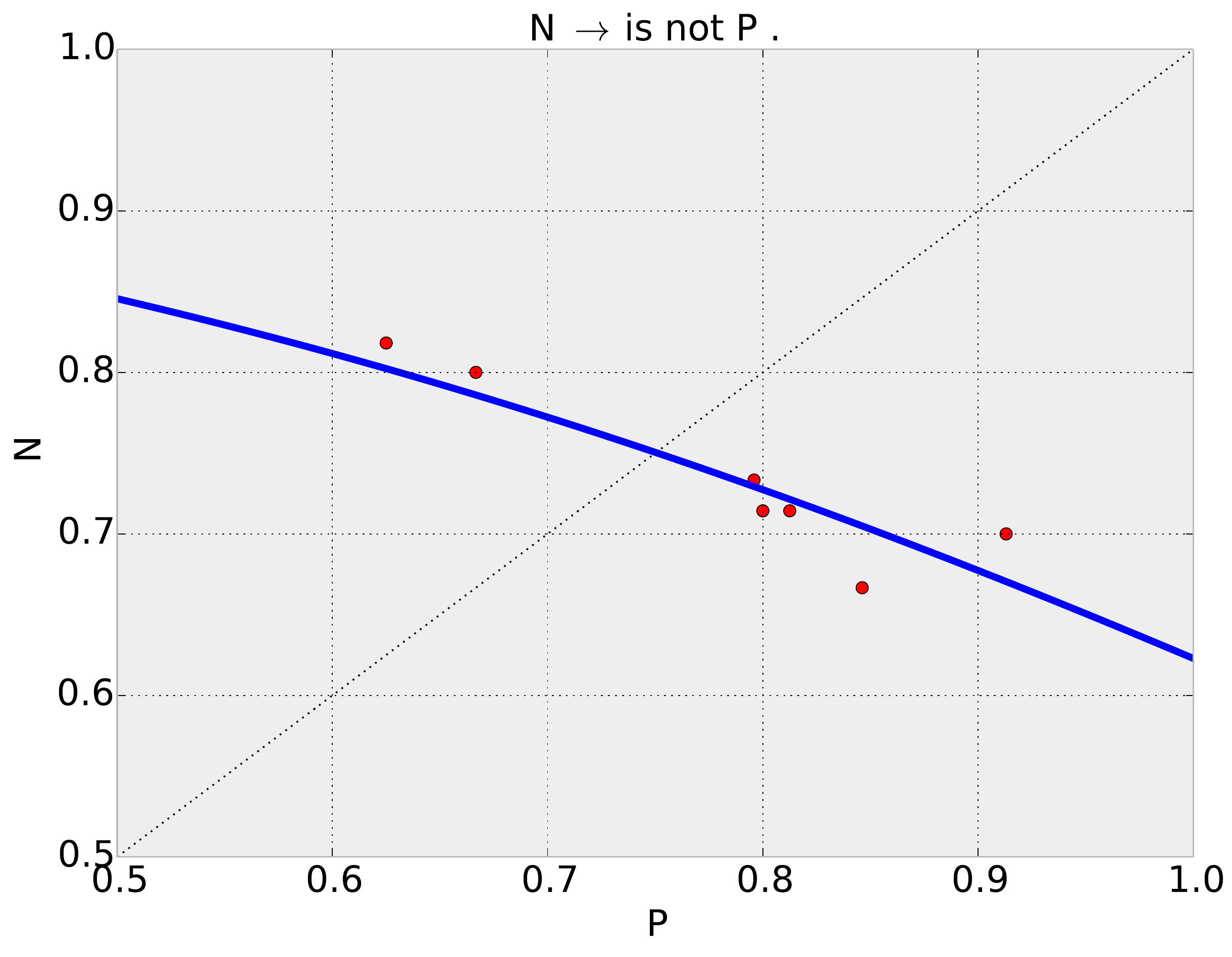}
}
\hfill
\subfloat[$P \rightarrow P~\mbox{actress}$\label{fig:experiment_computation_model:e}]{%
	\includegraphics[width=0.47\textwidth]{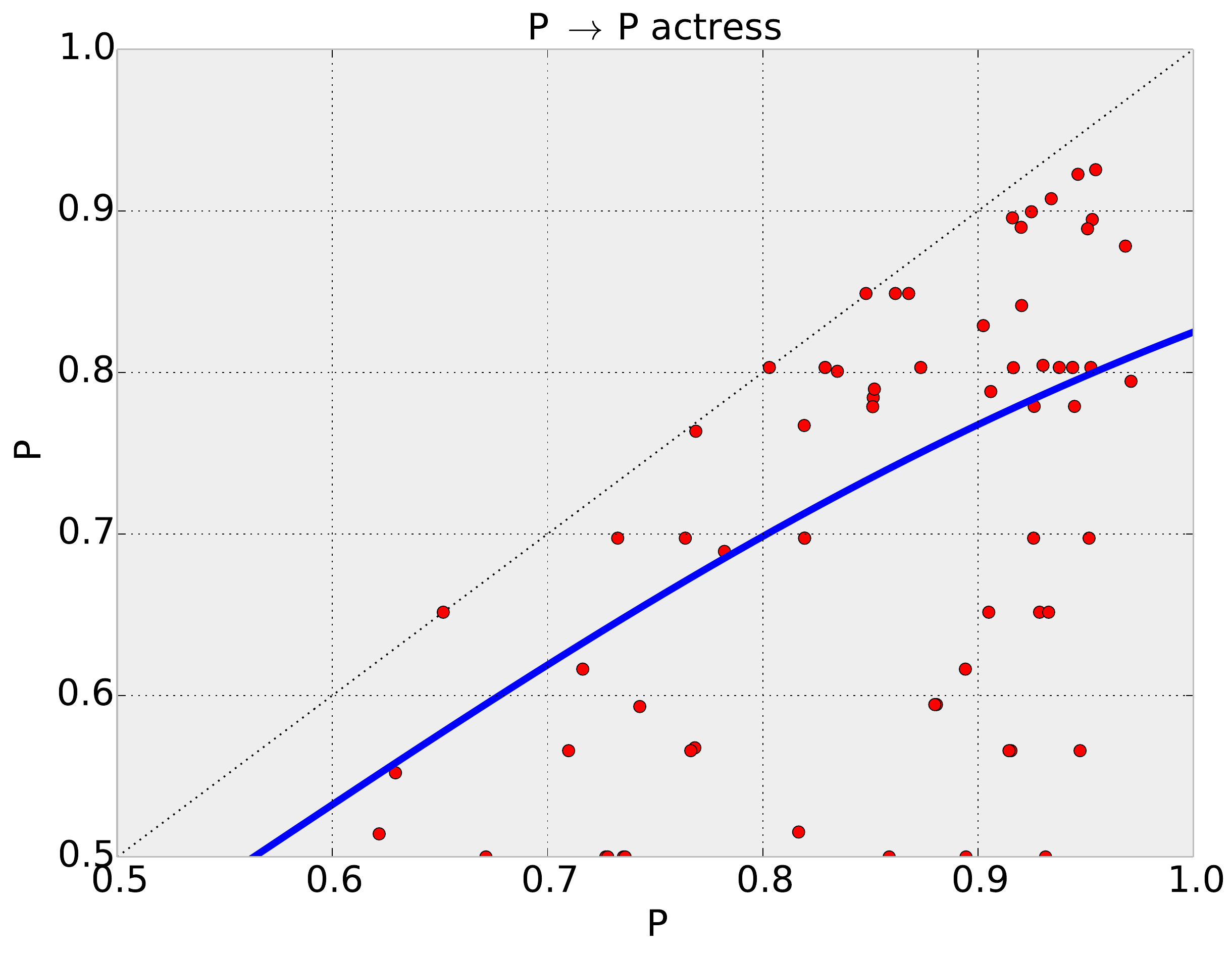}
}
\hfill
\subfloat[$N \rightarrow \mbox{a~bit~of}~N$\label{fig:experiment_computation_model:f}]{%
	\includegraphics[width=0.47\textwidth]{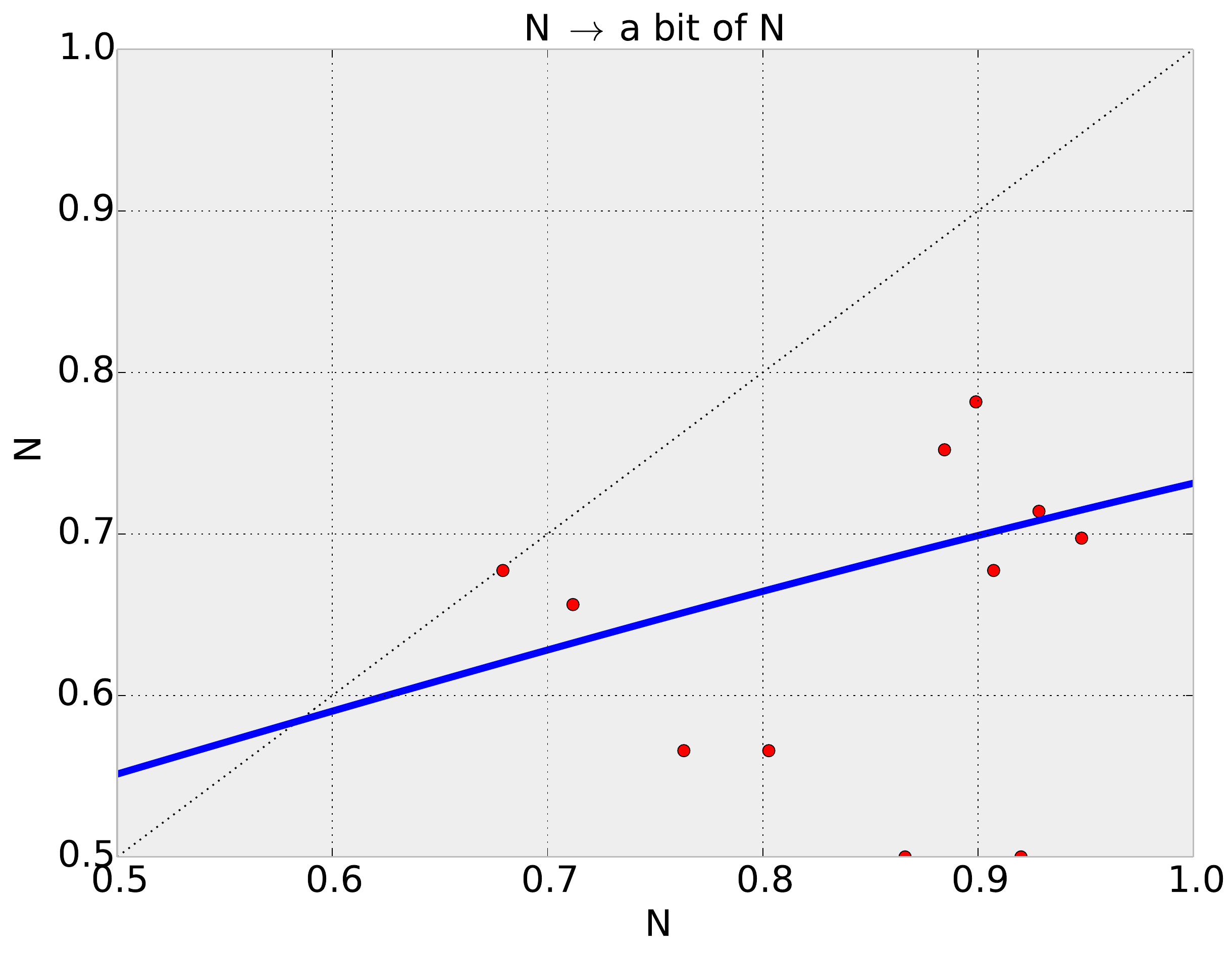}
}
\hfill
\subfloat[$P \rightarrow \mbox{lot~of}~P$\label{fig:experiment_computation_model:c}]{%
	\includegraphics[width=0.47\textwidth]{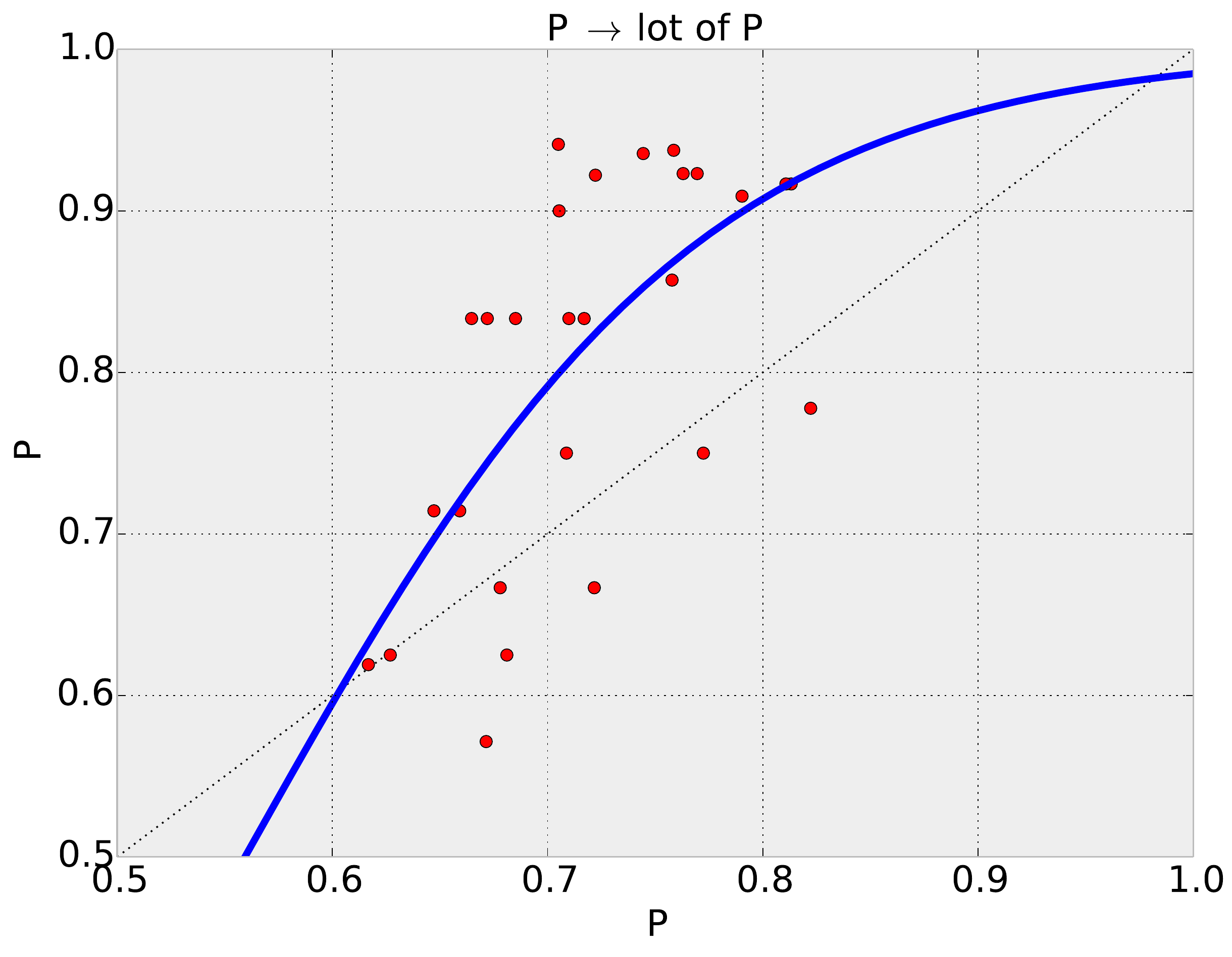}
}
\hfill
\subfloat[$N \rightarrow N~\mbox{terribly}$\label{fig:experiment_computation_model:d}]{%
	\includegraphics[width=0.47\textwidth]{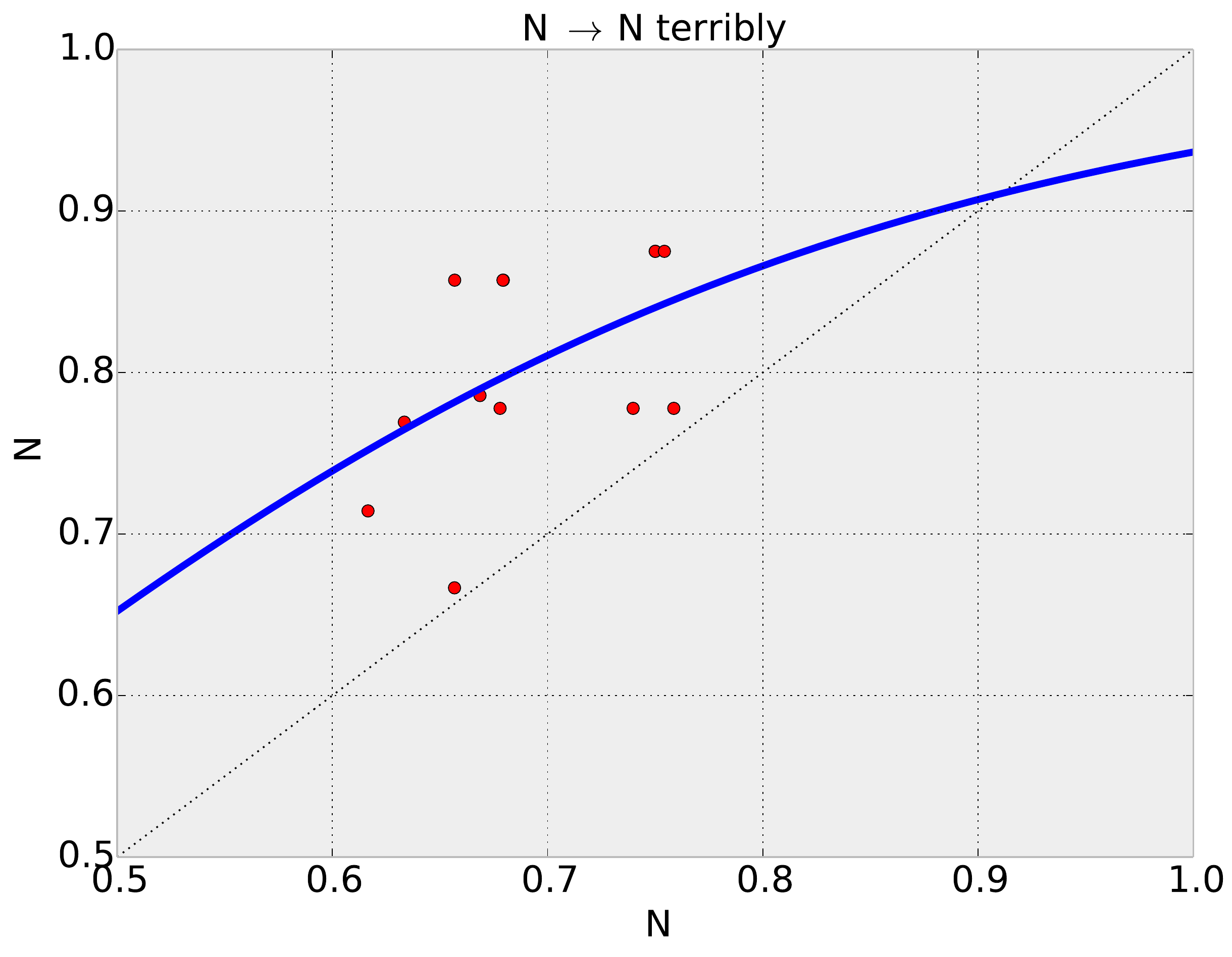}
}
\caption{Illustration of the polarity model for combination rules: (a)(b) Negation rule. (c)(d) Weaken rule. (e)(f) Strengthen rule. The labels of axes represent the corresponding polarity labels, the red points are the training instances, and the blue lines are the regression results for the polarity model.}
\label{fig:experiment_computation_model}
\end{center}
\end{figure}

In Figure~\ref{fig:experiment_computation_model}, we show the polarity model of some combination rules learned from the dataset RT-C.
The first two examples are negation rules. We find that both switch negation and shift negation exist in data, instead of using only one negation type in previous work~\cite{sauri2008factuality,Choi:2008:compostional,cl:lexicon}.
For the rule ``$N \rightarrow \mbox{i~do~not}~P$'', we find that it is a switch negation rule. This rule reverses the polarity and the corresponding polarity strength. For instance, the ``\textit{i do not like it very much}'' is more negative than the ``\textit{i do not like it}''.
As shown in Figure~\ref{fig:experiment_computation_model:b}, the ``$N \rightarrow \mbox{is~not}~P .$'' is a shift negation which reduces a fixed polarity strength to reverse the original polarity. Specifically, the ``\textit{is not good}'' is more negative than the ``\textit{is not great}'' as described in Section~\ref{sec:polarity_model}.
We have a similar conclusion for the next two weaken rules.
As illustrated in Figure~\ref{fig:experiment_computation_model:e}, the ``$P \rightarrow P~\mbox{actress}$'' describes one aspect of a movie, hence it is more likely to decrease the polarity intensity. We find that this rule is a fixed intensification rule which reduces the polarity probability by a fixed value.
The ``$N \rightarrow \mbox{a~bit~of}~N$'' is a percentage intensification rule, which scales polarity intensity by a percentage. It reduces more strength for stronger polarity.
The last two rules in Figure~\ref{fig:experiment_computation_model:c} and Figure~\ref{fig:experiment_computation_model:d} are strengthen rules. Both ``$P \rightarrow \mbox{lot~of}~P$'' and ``$N \rightarrow N~\mbox{terribly}$'' increase the polarity strength of the sub-fragments.
These cases indicate that it is necessary to learn how the context performs compositionality from data. In order to capture the compositionality for different rules, we define the polarity model and learn parameters for each rule.
This also agrees with the models of~\namecite{SocherEtAl2012:MVRNN} and~\namecite{dong2014adaptive}, which use multiple composition matrices to make compositions specific and improves over the recursive neural network which employs one composition matrix.

\section{Conclusion and Future Work} \label{sec:conclusion}
In this article, we propose a statistical parsing framework for sentence-level sentiment classification, which provides a novel approach to designing sentiment classifiers from a new perspective. It directly analyzes the sentiment structure of a sentence other than relying on syntactic parsing results as in existing literature. We show that complicated phenomena in sentiment analysis, such as negation, intensification, and contrast, can be handled the same as simple and straightforward sentiment expressions in a unified and probabilistic way.
We provide a formal model to represent the sentiment grammar built upon CFGs (Context-Free Grammars). The framework consists of: (1) a parsing model to analyze the sentiment structure of a sentence; (2) a polarity model to calculate sentiment strength and polarity for each text span in the parsing process; and (3) a ranking model to select the best parsing result from a list of candidate sentiment parse trees. We show that the sentiment parser can be trained from the examples of sentences annotated only with sentiment polarity labels but without using any syntactic or sentiment annotations within sentences.
We evaluate the proposed framework on standard sentiment classification datasets. The experimental results show the statistical sentiment parsing notably outperforms the baseline sentiment classification approaches.

We believe the work on statistical sentiment parsing can be advanced from many different perspectives. First, statistical parsing has been a well-established research field, in which many different grammars and parsing algorithms have been proposed in previously published literature. It will be a very interesting direction to apply and adjust more advanced models and algorithms from the syntactic parsing and the semantic parsing to our framework. We leave it as a line of future work.
Second, we can incorporate target and aspect information in the statistical sentiment parsing framework to facilitate the target-dependent and aspect-based sentiment analysis. Intuitively, this can be done by introducing semantic tags of targets and aspects as new non-terminals in the sentiment grammar and revising grammar rules accordingly. However, acquiring training data will be an even more challenging task as we need more fine-grained information.
Third, as the statistical sentiment parsing produces more fine-grained information (e.g., the basic sentiment expressions from the dictionary rules as well as the sentiment structure trees), we will have more opportunities to generate better opinion summaries.
Moreover, we are interested in jointly learning parameters of the polarity model and the parsing model from data.
Last but not the least, we are interested in investigating the domain adaptation which is a very important and challenging problem in sentiment analysis. Generally, we may need to learn domain-specific dictionary rules for different domains while we believe combination rules are mostly generic across different domains. This is also worth consideration for further study in future works.

\starttwocolumn

\begin{acknowledgments}
This research was partly supported by NSFC (Grant No. 61421003) and the fund of the State Key Lab of Software Development Environment (Grant No. SKLSDE-2015ZX-05).
We gratefully acknowledge helpful discussions with Dr. Nan Yang and the anonymous reviewers.
\end{acknowledgments}

\bibliographystyle{fullname}
\bibliography{ref}

\end{document}